\documentclass[10pt,twocolumn,letterpaper, table]{article}

\usepackage[pagenumbers]{cvpr}              

\usepackage{multirow}
\usepackage{adjustbox} 
\usepackage{booktabs}
\usepackage{siunitx}
\usepackage{amsmath}
\usepackage{stfloats}
\usepackage{float}
\usepackage{microtype}
\usepackage[accsupp]{axessibility}

\definecolor{rankone}{RGB}{227, 252, 227}
\definecolor{ranktwo}{RGB}{255, 249, 230}
\definecolor{rankthree}{RGB}{235, 245, 255}

\definecolor{cvprblue}{rgb}{0.21,0.49,0.74}
\usepackage[pagebackref,breaklinks,colorlinks,allcolors=cvprblue]{hyperref}

\def\confName{CVPR}
\def\confYear{2026}

\title{GEAR: GEometry-motion Alternating Refinement for Articulated Object Modeling with Gaussian Splatting}

\author{
  Jialin Li$^{1,2}$, Bin Fu$^{1,2}$, Ruiping Wang$^{1,2, \dag}$, Xilin Chen$^{1,2}$ \\
  $^1$Key Laboratory of AI Safety of Chinese Academy of Sciences (CAS), \\
Institute of Computing Technology, CAS, Beijing, 100190, China \\
$^2$University of Chinese Academy of Sciences, Beijing, 100049, China \\
  {\tt\small \{jialin.li, bin.fu\}@vipl.ict.ac.cn, \{wangruiping, xlchen\}@ict.ac.cn}
  }

\begin{document}
\maketitle
\let\thefootnote\relax\footnotetext{$\dag$ Corresponding author}
\begin{abstract}
High-fidelity interactive digital assets are essential for embodied intelligence and robotic interaction, yet articulated objects remain challenging to reconstruct due to their complex structures and coupled geometry-motion relationships. Existing methods suffer from instability in geometry-motion joint optimization, while their generalization remains limited on complex multi-joint or out-of-distribution objects. To address these challenges, we propose GEAR, an EM-style alternating optimization framework that jointly models geometry and motion as interdependent components within a Gaussian Splatting representation. GEAR treats part segmentation as a latent variable and joint motion parameters as explicit variables, alternately refining them for improved convergence and geometric-motion consistency. To enhance part segmentation quality without sacrificing generalization, we leverage a vanilla 2D segmentation model to provide multi-view part priors, and employ a weakly supervised constraint to regularize the latent variable. Experiments on multiple benchmarks and our newly constructed dataset GEAR-Multi demonstrate that GEAR achieves state-of-the-art results in geometric reconstruction and motion parameters estimation, particularly on complex articulated objects with multiple movable parts. Project Page:
\href{https://github.com/VIPL-VSU/GEAR}{https://github.com/VIPL-VSU/GEAR}.

\end{abstract}    
\section{Introduction}
\label{sec:intro}

In fields such as embodied AI~\cite{duan2022survey, liu2025aligning}, robotics~\cite{zhang2025twinor, hsu2023ditto,ma2023sim2real}, and virtual/augmented reality (VR/AR)~\cite{behravan2025transcending,venkatesan2021virtual}, high-fidelity, interactive digital assets are a key foundation for enabling agents to understand and interact with the real world.
Among various interactive objects, articulated objects (e.g., cabinet doors, laptops, scissors, etc.) are particularly common, with more complex structures and higher interaction difficulty.
However, due to the higher reconstruction difficulty, high-quality digital assets for articulated objects remain scarce, limiting the development of related research and applications~\cite{chen2024urdformer,sun2025arti,xiang2020sapien}.
Therefore, the study of efficient and high-fidelity articulated object modeling methods is of great importance.

\begin{figure}[t]
\centering
\includegraphics[width=1\linewidth]{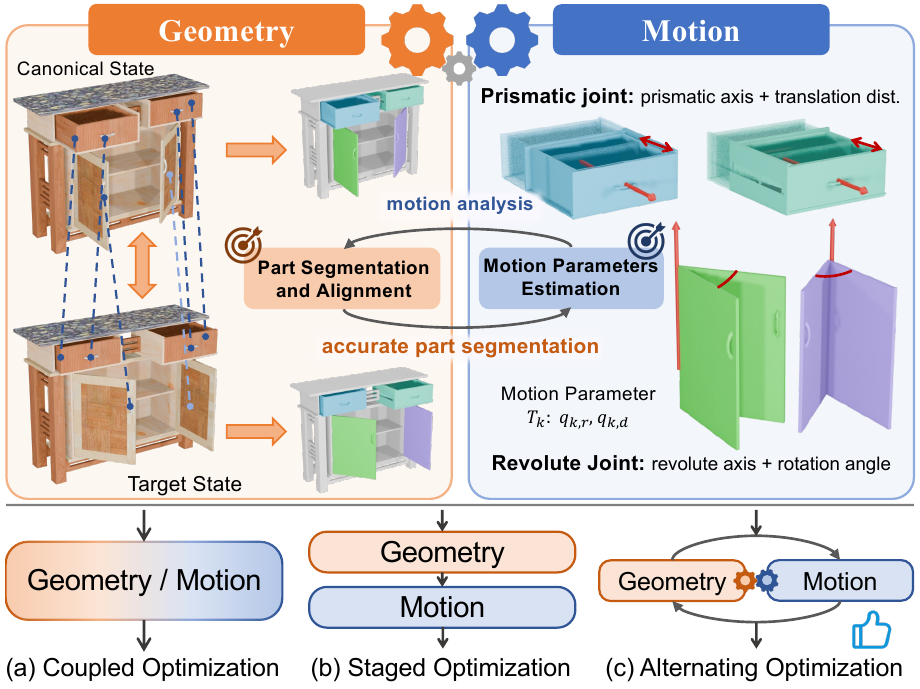}
\vspace{-1.6em}
\caption{Articulated object modeling involves coupled optimization of geometry and motion. Motion analysis plays a key role in part reconstruction, while joint motion parameter estimation relies on accurate part segmentation. Unlike coupled and staged methods, GEAR utilizes alternating optimization to improve stability.}
\label{fig:head}
\vspace{-1.6em}
\end{figure}

Compared to general rigid objects, articulated objects consist of multiple movable components, and their motion induces significant geometric deformation, resulting in diverse object states.
As such, the modeling of articulated objects requires reconstructing part-level geometric structures, and joint motion parameters from multi-view, multi-state observations, a highly complex and challenging process~\cite{artgs, Ditto, Paris,weng2024neural, GaussianArt}.
One of the most critical issues in this task is the coupled optimization between geometry and motion: accurate part segmentation often relies on the analysis of part motion behaviors across different states, while the estimation of motion parameters depends on reliable part segmentation results, as shown in \cref{fig:head}.
The mutual dependence of these two aspects makes the optimization process highly coupled, significantly increasing optimization uncertainty and computational complexity.

In recent years, early methods~\cite{weng2024neural,Paris,artgs,Reartgs} for high-fidelity modeling of articulated objects typically performed joint optimization of part segmentation and motion parameters, but faced significant instability.
To address this, subsequent works explored alternative strategies, such as staged decoupling~\cite{SplArt} or joint optimization followed by motion refinement~\cite{GaussianArt}, which improved stability but became highly dependent on the accuracy of the initial part segmentation.
However, existing initialization methods-whether unsupervised clustering for simple objects~\cite{artgs} or segmentation models (e.g., SAM~\cite{sam}) fine-tuned with in-distribution data, exhibit significant generalization issues when facing complex multi-joint or out-of-distribution data.

To tackle these challenges, we propose \textbf{GEAR} (\textbf{GE}ometry-motion \textbf{A}lternating \textbf{R}efinement), an alternating optimization framework where the object representation is based on Gaussian Splatting.
This framework draws inspiration from the Expectation-Maximization (EM) algorithm~\cite{moon1996expectation}, which is ideal for structure inference tasks with bidirectional dependencies.
In GEAR, we treat motion parameters as model parameters
and part segmentation as latent variables,
achieving stable convergence through alternating optimization.
To improve the quality and generalization of part segmentation, we introduce vanilla SAM during the segmentation optimization phase to provide weak supervision, enhancing GEAR’s stable and structure-aware segmentation for complex articulated objects.

Specifically, GEAR follows the EM paradigm and consists of three stages: initialization, E-step, and M-step, forming a coarse-to-fine modeling process.
In the initialization stage, we design a coarse reconstruction module
first obtains preliminary part segmentation and motion parameters, providing a stable starting point for subsequent optimization.
In the E-step, we update the part assignment probabilities for each Gaussian while fixing the motion parameters.
We leverage multi-view consistency constraints combined with semantic priors from 2D segmentation models (e.g., SAM~\cite{sam}) to furnish weak supervision for Gaussian mask optimization, yielding robust and consistent part segmentation across viewpoints.
In the M-step, we optimize the motion parameters for each joint while fixing the part assignments.
The E-step and M-step alternate, driving the part geometry and motion to iteratively converge to a physically consistent and high-fidelity reconstruction.

Our main contributions are summarized as follows:

\begin{itemize}
    \item We propose GEAR, an EM-based alternating optimization framework that effectively resolves the geometry-motion coupling, ensuring stable convergence for articulated object reconstruction.
    \item We introduce weak supervision from vanilla SAM to guide the E-step, achieving robust segmentation for complex objects with strong generalization.
    \item We construct a new dataset to address limitations in existing benchmarks. We demonstrate that GEAR outperforms existing state-of-the-art methods, excelling particularly in complex multi-joint objects.
\end{itemize}
\section{Related Work}
\label{sec:relatedwork}

\subsection{Articulated Object Modeling}

Articulated object modeling necessitates accurate part-level geometric reconstruction and kinematic parameter estimation. Early methods primarily rely on point clouds to predict part segmentation and motion \cite{yi2018deep,yan2019rpm,wang2019shape2motion,li2020category}. More recent approaches leverage single- or multi-state RGB images \cite{chen2024urdformer,gao2025partrm,geng2024sage,NEURIPS2023_3af8c40d,li2024locate,mandi2025realcode,wu2025predict,xia2025drawer,IAAO}, coupled with implicit neural representations \cite{oechsle2021unisurf,takikawa2021neural,NEURIPS2021_e41e164f} or 3D Gaussian Splatting (3DGS) \cite{kerbl20233d}, to jointly optimize geometry and articulation.

For motion estimation, existing methods broadly falls into two categories. \textbf{Prediction-based methods} learn articulation priors from large-scale 3D datasets \cite{heppert2023carto,wei2022self,kawana2021unsupervised,mu2021sdf,Ditto,ma2023sim2real,nie2022structure,hsu2023ditto,li2020category,tseng2022cla}, or leverage Foundation Models to directly infer kinematic parameters \cite{mandi2025realcode,le2025articulateanything}. However, acquiring high-quality annotations for articulated objects is notoriously difficult. The scarcity of diverse, real-world training data often restricts the zero-shot generalization capability of these data-driven approaches \cite{weng2024neural}.

The second category comprises \textbf{per-object optimization methods}, which directly fit articulated models to multi-state observations \cite{Paris,liu2023building,NEURIPS2024_d850b7e0,swaminathan2024leia,weng2024neural,artgs}. The primary bottlenecks for these approaches lie in accurately segmenting dynamic parts and resolving the severe coupling between geometric reconstruction and motion estimation, which often causes failures on complex multi-joint objects. To mitigate this, recent work like GaussianArt \cite{GaussianArt} relies on a fine-tuned Segment Anything Model (SAM) for precise initialization. However, requiring category-specific fine-tuning severely limits generalization to unseen domains. In contrast, our framework employs the vanilla SAM purely as a weak supervision signal, breaking the dependency loop through an alternating optimization strategy that effectively decouples geometry and motion.

\subsection{Dynamic Gaussian Splatting}

Recently, 3D Gaussian Splatting (3DGS) \cite{kerbl20233d} and its variants (e.g., 2DGS \cite{2dgs}) have revolutionized novel view synthesis across various 3D domains \cite{keetha2024splatam,wu2024hgs,jin2024gs,qian20243dgs,xie2024physgaussian,li2024robogsim,fu2025gs}. Benefiting from an explicit point-based representation, real-time rendering, and ease of editing, 3DGS has naturally become the mainstream representation for creating high-fidelity interactive digital assets that require both photorealistic appearance and distinct geometric structures.

To model moving scenes, Dynamic 3DGS approaches incorporate temporal dimensions \cite{luiten2024dynamic,sun20243dgstream,shaw2024swings} or continuous deformation fields \cite{yang2024deformable,liang2025gaufre,wu20244d,duisterhof2023md,lin2024gaussian,kratimenos2024dynmf} into the static 3DGS framework. However, these general dynamic methods fundamentally rely on dense, continuous multi-view video streams to track unconstrained motions. In contrast, articulated object modeling emphasizes inferring strict, physically constrained kinematic parameters (e.g., rotation axes and translation vectors) often from sparse, discrete observation states. Therefore, instead of learning a generic, unconstrained deformation field, our method explicitly parameterizes and optimizes rigid part kinematics within the Gaussian Splatting framework.
\section{Method}
\label{sec:method}
\subsection{Problem Statement}
\label{subsec:initialization}

Given multi-view RGB-D images $\mathcal{I}_s = \{(I_s^i, D_s^i, C_s^i)\}_{i=1}^N$ of an articulated object in two different states $s \in \{0, 1\}$, where $I$, $D$, and $C$ denote RGB images, depth maps, and camera parameters respectively, along with the known number of movable parts $\mathcal{K}$, the articulated object modeling task outputs the part segmentation $\mathcal{M}$ of $\mathcal{K}+1$ parts (including 1 static) and the joint motion parameters $\mathcal{T}$ of $\mathcal{K}$ movable parts.
Following \cite{GaussianArt}, we define the state with higher visibility (e.g., an open drawer) as the canonical state ($s=0$), and the state with lower visibility (e.g., a closed drawer) as the target state ($s=1$).

We employ 2D Gaussian Splatting~\cite{2dgs} as our fundamental representation, leveraging its accurate geometry estimation for articulated modeling. Each 2D Gaussian $g_i \in \mathcal{G}$ is defined as a 2D elliptical disk embedded in 3D space, parameterized by its position $p_i \in \mathbb{R}^3$, rotation matrix $r_i \in \mathbb{R}^{3\times 3}$, radial scale vector $\text{scale}_i\in \mathbb{R} ^2$, opacity $o_i \in \mathbb{R}$, and color $c_i \in \text{SH}(\cdot)$. By independently reconstructing from the two sets of input RGB-D images, we obtain Gaussian sets $\mathcal{G}_0$ and $\mathcal{G}_1$ for the two states, respectively trained via the vanilla Gaussian Splatting pipeline.

To explicitly model part segmentation at the Gaussian level, we assign a learnable mask vector $m_i \in \mathbb{R}^{\mathcal{K}+1}$ for each Gaussian $g_i \in \mathcal{G}_0$ in the canonical state, 
The collection of all mask vectors $\mathcal{M} = \{m_i\}$ constitutes the optimization variable for part segmentation. To model part-level motion, 
we learn a set of transformation matrices $\mathcal{T} = \{T_k  \in SE(3)\}_{k=0}^{\mathcal{K}}$, where $T_0$ is fixed as the identity matrix for the static part, and $T_k$ ($k \ge 1$) denotes the rigid body transformation of part $k$ from state $s=0$ to $s=1$.

\begin{figure*}[t]
\centering
\includegraphics[width=1\linewidth]{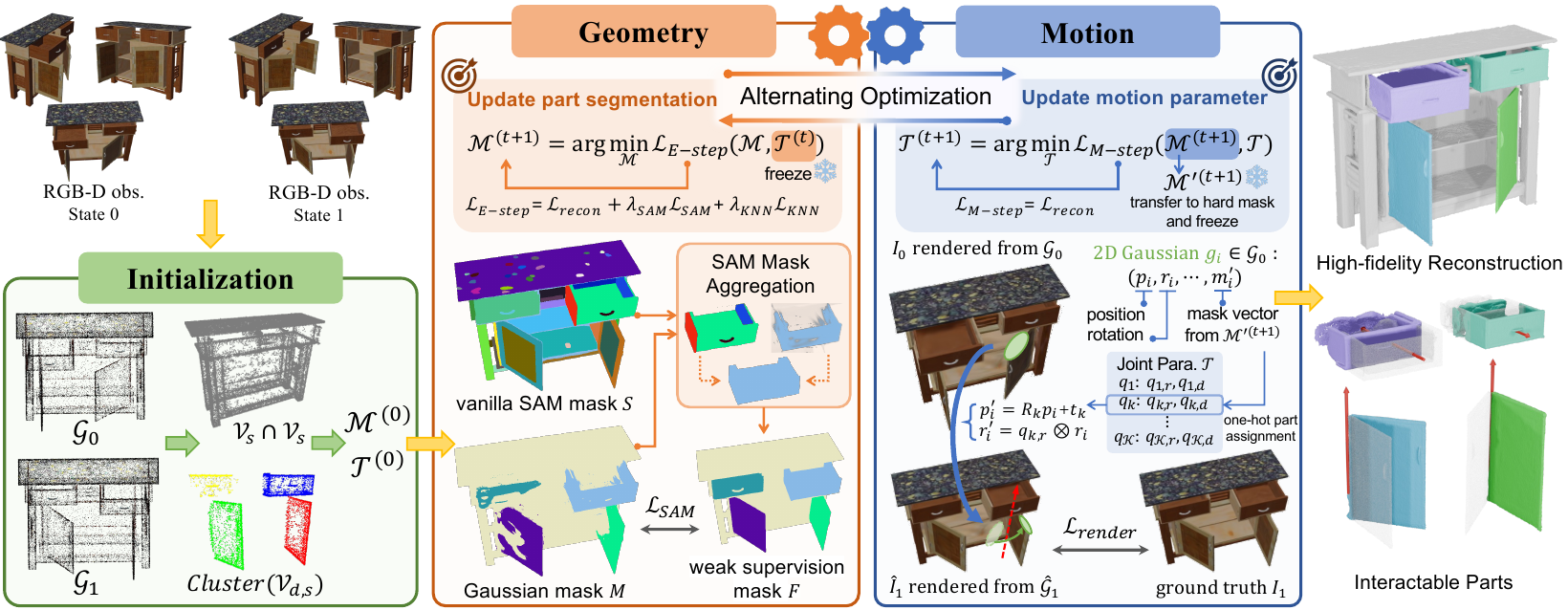}
\vspace{-1.6em}
\caption{\textbf{GEAR} is an EM-style framework with three modules: Initialization, geometry modeling, and motion modeling. The key idea is the alternating optimization of geometry and motion, which enhances the stability and performance of articulated object modeling.}
\label{fig:main}
\vspace{-1.6em}
\end{figure*}

\subsection{Geometry-Motion Alternating Optimization}
\label{subsec:em_optimization}

Previous work \cite{artgs, GaussianArt, Articulatedgs} formulate the optimization objective to learn $\mathcal{M}$ and $\mathcal{T}$ in the canonical state, such that the geometry $\mathcal{G}_0$ of the canonical state, after being transformed by the part segmentation mask $\mathcal{M}$ and transformation matrices $\mathcal{T}$, yields renderings that closely align with the observations $\mathcal{I}_1$ of the target state.
Based on the explicit geometric representation of GS, for $g_i \in \mathcal{G}_0$, this transformation and rendering process can be expressed as:
\begin{equation}
\hat{\mathcal{I}}_1 = \boldsymbol{\pi}\left(\left\{ \left(\sum_{k=0}^{\mathcal{K}} m_{i,k} T_k\right) \circ g_i \;\middle|\; g_i \in \mathcal{G}_0 \right\}\right),
\label{objective}
\end{equation}
where $m_{i,k}$ represents the probability of the Gaussian $g_i$ belonging to the $k$-th part ($k=0$ denotes the static part), $\circ$ denotes the application of spatial transformation to the Gaussian attributes, and $\boldsymbol{\pi}(\cdot)$ represents the differentiable rendering process. It reveals that both part segmentation $\mathcal{M}$ and motion parameters $\mathcal{T}$ simultaneously govern the rendering and optimization process.

Moreover, as illustrated in \cref{fig:head}, these two optimization objectives are highly coupled. Direct joint optimization suffers from ambiguity in error attribution: the optimizer tends to minimize the rendering loss by improperly distorting geometric structures to compensate for incorrect motion estimates (or vice versa), trapping the model in non-physical local minima. Conversely, staged optimization relies on a one-way pipeline that lacks a feedback mechanism, preventing the subsequent motion modeling stage from rectifying incorrect initial part assignments.

To resolve this dependency loop, we draw inspiration from the Expectation-Maximization (EM) algorithm and propose \textbf{GEAR} (\textbf{GE}ometry-motion \textbf{A}lternating \textbf{R}efinement), an alternating optimization framework.

As shown in \cref{fig:main}, GEAR decomposes the joint optimization of geometry and motion into two alternating steps. Since the Gaussian mask $\mathcal{M}$ describes finer structural partitions and indirectly determines the optimization direction and convergence quality of motion parameters $\mathcal{T}$, we treat $\mathcal{M}$ as the latent variable and $\mathcal{T}$ as model parameters, together forming complementary representations of geometry and motion.
The overall optimization process comprises three stages: \textbf{Initialization}, \textbf{E-step}, and \textbf{M-step}. The Initialization stage provides robust initial estimates $\mathcal{M}^{(0)}$ and $\mathcal{T}^{(0)}$ through a coarse reconstruction module; the E-step and M-step alternately refine the part segmentation $\mathcal{M}^{(t)}$ and motion parameters $\mathcal{T}^{(t)}$.

\textbf{Initialization: Coarse Reconstruction Module.} 
Direct optimization of $\mathcal{M}$ and $\mathcal{T}$ from scratch is prone to instability.
Therefore, we design a voxel-based coarse reconstruction module to provide robust initial estimates $\mathcal{M}^{(0)}$ and $\mathcal{T}^{(0)}$.
We first utilize the geometric distributions of Gaussian sets $\mathcal{G}_0$ and $\mathcal{G}_1$ to construct corresponding voxel occupancy sets $\mathcal{V}_0$ and $\mathcal{V}_1$ at voxel scale $v$. To identify dynamic regions, we define a voxel difference function:
\begin{equation}
\Phi(\mathcal{V}_s, \mathcal{V}_{\bar{s}}) = \mathcal{V}_s \setminus \mathcal{D}(\mathcal{V}_s \cap \mathcal{V}_{\bar{s}}),
\end{equation}
where $\bar{s}$ denotes the opposite state of $s$, $\mathcal{V}_s \cap \mathcal{V}_{\bar{s}}$ represents the static voxel set, and $\mathcal{D}(\cdot)$ is a morphological dilation operation to smooth noisy boundaries. The output of $\Phi$ is the dynamic candidate voxel set $\mathcal{V}_{d,s}$ for state $s$.

We perform connected component clustering $\mathcal{V}_{c,s} = \text{Cluster}(\mathcal{V}_{d,s})$ on $\mathcal{V}_{d,s}$ and select the top $\mathcal{K}$ largest voxel clusters as initial parts. Gaussians belonging to these clusters are initialized as dynamic Gaussians with distinct labels, while the remaining Gaussians are marked as static, yielding the initial mask $\mathcal{M}^{(0)}$.
Furthermore, to obtain reasonable initial motion parameters, we estimate $\mathcal{T}^{(0)}$ via a registration method that maximizes voxel overlap.
Finally, we employ the EM-style alternating optimization framework to jointly refine based on this coarse solution. The visualization results of this module are provided in Supplementary Sec. A.

\textbf{E-Step: Geometry Modeling}.
At iteration $t$, we fix the current motion parameters $\mathcal{T}^{(t)}$ and optimize the part segmentation $\mathcal{M}$ to minimize the rendering error:
\begin{equation}
\mathcal{M}^{(t+1)} = \arg\min_{\mathcal{M}} \mathcal{L}_{\mathrm{E\text{-}step}}(\mathcal{M}, \mathcal{T}^{(t)}).
\end{equation}

\textbf{M-Step: Motion Modeling}.
With the updated part segmentation $\mathcal{M}^{(t+1)}$ fixed, we further optimize the rigid motion parameters of the parts:
\begin{equation}
\mathcal{T}^{(t+1)} = \arg\min_{\mathcal{T}} \mathcal{L}_{\mathrm{M\text{-}step}}(\mathcal{M}^{(t+1)}, \mathcal{T}).
\end{equation}

The methodological details of geometry modeling and motion modeling are described in \cref{subsec:geometry_modeling} and \cref{subsec:motion_modeling}.

\subsection{Geometry Modeling}
\label{subsec:geometry_modeling}

In the geometry modeling phase, we fix the current motion parameters $\mathcal{T}$ and optimize the part segmentation $\mathcal{M} = \{m_i\}$. While fixing motion parameters reduces the difficulty of part segmentation optimization, for articulated objects with complex multi-joint structures, we still require stronger part segmentation priors. Given SAM's powerful image segmentation capability and its ability to provide masks with sharp boundaries, we leverage its segmentation masks $S$ to supervise the Gaussian part segmentation $\mathcal{M}$. However, SAM-generated masks vary in granularity and lack multi-view consistency, making them difficult to align with the part-level, multi-view consistent segmentation required in geometric optimization.

\begin{figure}[t]
\centering
\includegraphics[width=1\linewidth]{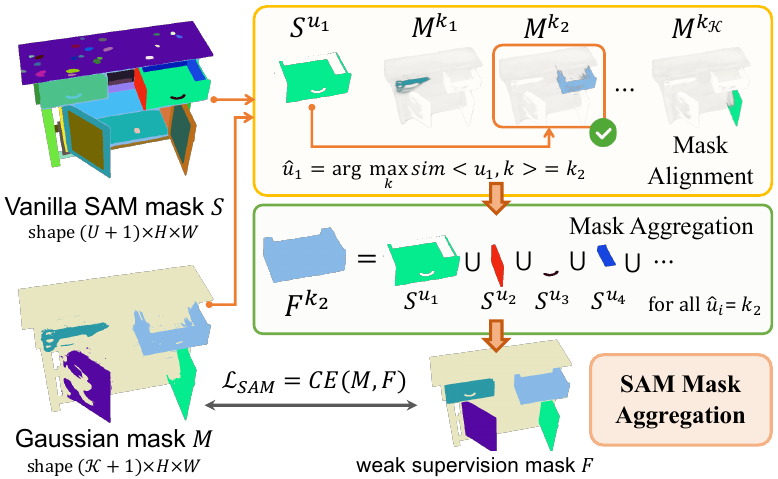}
\vspace{-1.6em}
\caption{SAM Mask Aggregation module. Each fine-grained candidate region $S^u$ from SAM is assigned to the Gaussian part $M^k$ with the highest spatial overlap. All SAM regions assigned to the same part $k$ are then aggregated to form a coherent mask $F^k$, serving as a weak supervision signal to refine $M^k$.}
\label{fig:sam}
\vspace{-1.6em}
\end{figure}

To address this challenge, we introduce the \textbf{SAM Mask Aggregation module}, as detailed in \cref{fig:sam}. Specifically, we employ SAM in its default automatic mode to extract fine-grained candidate regions $S = \{S^u\}_{u=0}^{U}$ from the images, where $S^u \in \{0,1\}^{H \times W}$ and $U$ denotes the number of candidate fine-grained SAM regions. This module employs a voting-based strategy to aggregate these fine-grained masks into a part-level consistent segmentation, providing weak supervision signals to guide the optimization of $\mathcal{M}$.

First, to enable differentiable optimization of $\mathcal{M}$ while achieving multi-view consistent segmentation, we design a soft probability-based segmentation map rendering method. This method generates soft Gaussian part segmentation maps $M = \{M^k\}_{k=0}^{\mathcal{K}}$ with $\mathcal{K}+1$ channels, and each map $M^k \in [0,1]^{H \times W}$ matches the dimensions of the input images. For a given pixel $x$ on the image, the segmentation map for part $k$ is expressed probabilistically as:
\begin{equation}
M^k(x) = \sum_{i=1}^{N} m_{i,k} \alpha_i \hat{g}_i(x) \prod_{j=1}^{i-1} (1 - \alpha_j \hat{g}_j(x)),
\end{equation}
where $\hat{g}_i(x)$ denotes the 2D projection value of Gaussian $g_i$ at pixel $x$, and $\alpha_i$ is its learned opacity parameter.

To establish the correspondence between the SAM priors $S$ and our rendered parts $M$, we compute the spatial overlap similarity between each SAM region $u$ and each part $k$:

\begin{equation}
\text{sim}\langle u,k \rangle = \sum S^u \odot M^k,
\end{equation}
where $\odot$ denotes element-wise multiplication and $\sum$ denotes the summation over all elements of the resulting matrix.
Leveraging this similarity, we assign each region $u$ in the SAM mask to the Gaussian part with the highest similarity, denoted as $\hat{u}= \arg\max_k (\text{sim}\langle u,k \rangle) \in \{0,1,\cdots,\mathcal{K}\}$.

Finally, based on the region-to-part assignments, we construct aggregated binary maps $F^k = \bigcup_{\hat{u}=k} S^u$, where $\cup$ denotes the element-wise logical OR operation, and $F^k \in \{0,1\}^{H \times W}$. Given $F = \{F^k\}_{k=0}^{\mathcal{K}}$, we use it as a weak supervision signal to refine the rendered part segmentation maps $M$ via a cross-entropy loss:

\begin{equation}
\mathcal{L}_{\text{SAM}} = \mathrm{CE}(M, F).
\end{equation}

Additionally, to prevent unstable diffusion of masks in 3D space and maintain continuity of masks within the same part, we introduce a KNN-based mask clustering consistency loss inspired by \cite{GaussianArt}:
\begin{equation}
\mathcal{L}_{\text{KNN}} = \frac{1}{N|\mathcal{N}(g_i)|} \sum_i \sum_{j \in \mathcal{N}(g_i)} \|m_i - m_j\|_2^2,
\end{equation}
where $N$ is the total number of Gaussians involved in the loss computation, $\mathcal{N}(g_i)$ denotes the nearest neighbors of $g_i$, and $|\mathcal{N}(g_i)|$ is the number of nearest neighbors. This constraint encourages similar mask representations in local neighborhoods, forming semantically consistent part clusters in 3D space.

As formulated in \cref{objective}, to optimize $\mathcal{M}$ and $\mathcal{T}$, we transform the geometry $\mathcal{G}_0$ of the canonical state $s=0$ to the target state $s=1$, and use the loss $\mathcal{L}_{\text{render}}^{(1)}$ between the rendered images and the target state observations to provide supervision.
Additionally, to prevent the model from forgetting the geometric features of the original state while approximating the target state, we introduce a self-rendering loss $\mathcal{L}_{\text{render}}^{(0)}$ supervised by state $s=0$.
The losses for the two states, $s=0$ and $s=1$, can be unified as follows:
\begin{equation}
\mathcal{L}_{\text{render}}^{(s)} 
= (1-\lambda_{\text{D-SSIM}})\,\mathcal{L}_{s}
+ \lambda_{\text{D-SSIM}}\,\mathcal{L}_{\text{D-SSIM}}
+ \lambda_{D}\,\mathcal{L}_{D},
\label{eq:render_loss}
\end{equation}
where $\mathcal{L}_{s}=\|\hat{I}_s - I_s\|_1$ denotes the image reconstruction loss,  $\mathcal{L}_{\text{D-SSIM}}=1-\text{SSIM}(\hat{I}_s,I_s)$ measures structural similarity, and $\mathcal{L}_{D}=\mathbb{E}[\log(1 + |\hat{D}_s - D_s|)]$ enforces depth consistency as well as mitigate large errors.
Here, $I_s$ and $D_s$ represent the ground truth images, while $\hat{I}_s$ and $\hat{D}_s$ are rendered images from $\hat{\mathcal{G}}_1$ (transformed from $\mathcal{G}_0$) or $\mathcal{G}_0$.

To summarize, the E-step optimizes both rendering quality and weakly supervised segmentation consistency, the final geometry modeling loss is defined as:
\begin{equation}
\mathcal{L}_{\text{E-step}} = \mathcal{L}_{\text{render}}^{(0)} + \mathcal{L}_{\text{render}}^{(1)} + \lambda_{\text{SAM}} \mathcal{L}_{\text{SAM}} + \lambda_{\text{KNN}} \mathcal{L}_{\text{KNN}}.
\end{equation}

\subsection{Motion Modeling}
\label{subsec:motion_modeling}

In the motion modeling phase, we optimize the motion parameters $\mathcal{T} = \{T_k\}$ while fixing the part segmentation $\mathcal{M}$. To ensure differentiability and maintain a unified structure for rotation and translation, following~\cite{artgs}, we adopt dual quaternions to uniformly model the rigid body motion.

For part $k$, its motion parameters are represented as $T_k = (q_{k,r}, q_{k,d})$, where $q_{k,r}$ and $q_{k,d}$ are the real and dual parts denoting the rotation and translation components, respectively. Inheriting the soft assignment probabilities $m_{i,k}$ from the geometry modeling phase, the continuous rigid transformation of a Gaussian $g_i$ can be generally formulated as a weighted blending:
\begin{equation}
q_i = \left( \sum_{k} m_{i,k} q_{k,r}, \sum_{k} m_{i,k} q_{k,d} \right).
\end{equation}

While this soft blending formulation ensures smooth gradient flow during the E-step's mask optimization, directly optimizing motion parameters under soft weights in the M-step often leads to non-physical geometric distortions. Therefore, strictly during the M-step, we enforce a deterministic hard assignment to guarantee absolute rigid-body constraints and promote sharp geometric boundaries. We binarize the influence by assigning each Gaussian exclusively to its most probable part, denoted as $\hat{k}=\text{argmax}_k{(m_{i,k})}$. This yields the exact pose transformation of the Gaussian $g_i$ from state $s=0$ to $s=1$:
\begin{equation}
p_i' = R_{\hat{k}} p_i + t_{\hat{k}}, \quad r_i' = R_{\hat{k}} r_i,
\end{equation}
where $(R_{\hat{k}}, t_{\hat{k}})$ are the rotation matrix and translation vector derived from the dual quaternion $q_{\hat{k}}$, and $\otimes$ represents quaternion multiplication. This process ensures continuous deformation of each Gaussian under strict part-level motion.

The final motion modeling loss is defined as:
\begin{equation}
\mathcal{L}_{\text{M-step}} = \mathcal{L}_{\text{render}}^{(0)} + \mathcal{L}_{\text{render}}^{(1)}.
\end{equation}

By alternating refinement between the E-step and M-step, GEAR decomposes the highly coupled geometry-motion optimization problem into two more stable subproblems, thereby establishing an iterative optimization loop that ensures both geometric and motion consistency.

Finally, to achieve accurate joint type estimation and improve parameter estimation accuracy, we extract the rotation angle $\theta_k$ from the part-level rotation quaternion $q_{k,r}$ at specific iteration steps, and classify parts with $\theta_k$ below a threshold $\epsilon$ as prismatic joints. For parts identified as prismatic joints, we fix their rotation part $q_{k,r}$ to the identity quaternion and only optimize the translation component $q_{k,d}$, thereby obtaining more precise estimates.

\section{Experiments}
\label{sec:experiment}

\begin{table*}[t]
 \centering
    \caption{  Quantitative comparison on the PARIS dataset~\cite{Paris}, averaged over 10 trials for both simulated and real objects with higher visibility. 
  Lower ($\downarrow$) is better for all metrics. 
  We highlight the best (\textbf{Bold} with \colorbox{rankone!90}{Top-1}) and second-best (\colorbox{ranktwo!90}{Top-2}) results. 
  Baseline results for Ditto, PARIS, DTA, and ArtGS are sourced from~\cite{artgs}.}
  \vspace{-0.6em}
 \resizebox{\textwidth}{!}{
 \begin{tabular}{l cccccc cccccc}
   \toprule
   \multirow{2}{*}{\textbf{Method}} & \multicolumn{6}{c}{\textbf{Simulated  (Average)}} & \multicolumn{6}{c}{\textbf{Real (Average)}} \\
   \cmidrule(lr){2-7} \cmidrule(lr){8-13}
   & Axis Ang & Axis Pos & Geo Dist & CD-s & CD-m & CD-w & Axis Ang & Axis Pos & Geo Dist & CD-s & CD-m & CD-w \\
    \midrule
    Ditto & 46.22 & 2.11 & 39.87 & 18.94 & 42.20 & 7.12 & 3.80 & 1.84 & 4.41 & 31.55 & 35.48 & 10.29 \\
    PARIS & 6.23 & 1.04 & 41.71 & 7.18 & 39.76 & 5.47 & 22.39 & \cellcolor{rankone!90}\textbf{0.34} & 1.36 & 48.56 & 455.24 & 43.17 \\
    DTA & \cellcolor{ranktwo!90}0.13 & \cellcolor{ranktwo!90}0.02 & \cellcolor{ranktwo!90}0.13 & \cellcolor{ranktwo!90}2.45 & 1.66 & \cellcolor{rankone!90}\textbf{1.79} & 7.86 & 0.59 & \cellcolor{ranktwo!90}1.00 & 6.67 & 15.95 & 5.53 \\
    ArtGS & \cellcolor{rankone!90}\cellcolor{rankone!90}\textbf{0.02} & \cellcolor{rankone!90}\textbf{0.00} & \cellcolor{rankone!90}\textbf{0.02} & 2.63 & \cellcolor{rankone!90}\textbf{0.67} & 2.15 & \cellcolor{rankone!90}\cellcolor{rankone!90}\textbf{2.78} & 0.47 & \cellcolor{rankone!90}\textbf{0.99} & \cellcolor{rankone!90}\textbf{2.29} & \cellcolor{rankone!90}\textbf{3.47} & \cellcolor{rankone!90}\textbf{2.26} \\
    Ours & \cellcolor{rankone!90}\textbf{0.02} & \cellcolor{rankone!90}\textbf{0.00} & \cellcolor{rankone!90}\textbf{0.02} & \cellcolor{rankone!90}\textbf{1.99} & \cellcolor{ranktwo!90}0.70 & \cellcolor{ranktwo!90}1.87 & \cellcolor{ranktwo!90}3.50 & \cellcolor{ranktwo!90}0.38 & 1.18 & \cellcolor{ranktwo!90}2.60 & \cellcolor{ranktwo!90}8.35 & \cellcolor{ranktwo!90}2.99 \\
    \bottomrule
  \end{tabular}
  } 
  \vspace{-1.5em}

    \label{tab:paris}
\end{table*}

\subsection{Experimental Setup}

\noindent\textbf{Datasets \& Baselines.} We evaluate GEAR on three datasets. PARIS~\cite{Paris} contains 12 single-joint objects (10 synthetic, 2 real-world). ArtGS-Multi~\cite{artgs} includes 5 multi-joint objects. To comprehensively evaluate complex structures, we construct \textbf{GEAR-Multi}, comprising 10 diverse multi-joint objects across 10 categories sourced from the PM dataset~\cite{xiang2020sapien}. 
For baselines, we compare against Ditto~\cite{Ditto}, PARIS, DTA~\cite{weng2024neural}, and ArtGS~\cite{artgs} on the PARIS dataset. On the more challenging multi-joint datasets (ArtGS-Multi and GEAR-Multi), we compare primarily with the recent state-of-the-art ArtGS.

\noindent\textbf{Metrics.} For geometric reconstruction, we uniformly sample 10K points on the reconstructed and ground-truth meshes to compute the Chamfer Distance for the whole object (CD-w), static parts (CD-s), and movable parts (CD-m). For motion estimation, we evaluate joint axis accuracy using the angular error (Axis Ang) between 3D directional vectors, and the spatial distance (Axis Pos) for revolute axes. Finally, we compute the articulation displacement error (Geo Dist) to assess the state transition accuracy, which measures the rotation or translation deviations from state 0 to state 1.

\begin{table}[h]
\centering
\caption{Quantitative comparison on the ArtGS-Multi dataset~\cite{artgs}. Numbers in parentheses (e.g., Table (4)) indicate the count of dynamic parts.
Results are reported as the mean over 3 trials. 
Lower ($\downarrow$) is better for all metrics, \textbf{Bold} indicates the best results. 
Axis Pos. is omitted for ``Table (4)" which contains only prismatic joints. 
Baseline results for DTA and ArtGS are sourced from~\cite{artgs}.}
\vspace{-0.3em}
\resizebox{\columnwidth}{!}{
\begin{tabular}{c l c c c c c c}
\hline
Object & Method & Axis Ang & Axis Pos & Geo Dist & CD-s & CD-m & CD-w \\ \hline
\multirow{3}{*}{Table (4)} 
 & DTA & 24.35 & - & 0.12 & 0.59 & 104.38 & 0.55 \\
 & ArtGS & 1.16 & - & \textbf{0.00} & 0.74 & 3.53 & 0.74 \\
 & Ours & \textbf{0.08} & - & \textbf{0.00} & \textbf{0.47} & \textbf{0.23} & \textbf{0.51}\\ \hline
\multirow{3}{*}{Table (5)} 
 & DTA & 20.62 & 4.2 & 30.8 & 1.39 & 230.38 & 1.00 \\
 & ArtGS & \textbf{0.04} & \textbf{0.00} & \textbf{0.01} & 1.22 & 3.09 & 1.16 \\
 & Ours & \textbf{0.04} & \textbf{0.00} & 0.02 & \textbf{0.77} & \textbf{0.26} & \textbf{0.98} \\ \hline
\multirow{3}{*}{Storage (4)} 
 & DTA & 51.18 & 2.44 & 43.77 & 5.74 & 246.63 & 0.88 \\
 & ArtGS & \textbf{0.02} & \textbf{0.00} & \textbf{0.03} & 0.75 & 0.13 & 0.88 \\
 & Ours & \textbf{0.02} & 0.01 & 0.04 & \textbf{0.62} & \textbf{0.10} & \textbf{0.75} \\ \hline
\multirow{3}{*}{Storage (7)} 
 & DTA & 19.07 & 0.31 & 10.67 & 0.82 & 476.91 & 0.71 \\
 & ArtGS & 0.14 & 0.02 & 0.62 & 0.67 & 3.70 & 0.70 \\
 & Ours & \textbf{0.07} & \textbf{0.01} & \textbf{0.10} & \textbf{0.61} & \textbf{0.32} & \textbf{0.60} \\ \hline
\multirow{3}{*}{Oven (4)} 
 & DTA & 17.83 & 6.51 & 31.80 & 1.17 & 359.16 & 1.01 \\
 & ArtGS & 0.04 & 0.01 & 0.23 & 1.08 & 0.25 & 1.03 \\
 & Ours & \textbf{0.02} & \textbf{0.00} & \textbf{0.04} & \textbf{0.68} & \textbf{0.13} & \textbf{0.66} \\ \hline
\multirow{3}{*}{\textbf{Average}} 
 & DTA & 26.61 & 3.37 & 23.43 & 1.94 & 283.49 & 0.83 \\
 & ArtGS & 0.28 & \textbf{0.01} & 0.18 & 0.89 & 2.14 & 0.90 \\
 & Ours & \textbf{0.05} & \textbf{0.01} & \textbf{0.04} & \textbf{0.63} & \textbf{0.21} & \textbf{0.70} \\ \hline
\end{tabular}
}
\vspace{-0.9em}

\label{tab:artgs}
\end{table}

\noindent\textbf{Implementation Details}.
To obtain accurate canonical Gaussian representations $\mathcal{G}_0$ and $\mathcal{G}_1$ and ensure stable convergence, 
we first perform 30k iterations of initialization to establish stable Gaussians, followed by 30k iterations of geometry–motion alternating optimization.
During training, both the E-step and M-step are executed for 500 iterations alternately, where we supervise reconstruction quality through two complementary rendering losses.
While our full optimization schedule prioritizes maximum stability for complex multi-joint objects, the EM framework intrinsically converges fast. When scaled down to a reduced schedule matching the baseline's iterations, our method maintains superior accuracy with even less runtime ($\sim$11 minutes per object). Detailed efficiency analyses and hyperparameter robustness are provided in Supplementary Material.

\subsection{Results}
\label{sec-results}

\begin{table*}[t]
\centering
\vspace{-0.6em}
\caption{Quantitative comparison on our GEAR-Multi dataset over 3 trials. Lower ($\downarrow$) is better for all metrics, \textbf{Bold} indicates the best results. Axis Pos is omitted for ``Clock" which contains only prismatic joints.}
\vspace{-0.6em}
\renewcommand{\arraystretch}{1.1}
\setlength{\tabcolsep}{2.5pt}
\begin{adjustbox}{width=\textwidth}
\begin{tabular}{c l | c c c c c c c c c c c}
\hline

 & & Box (5) & Bucket (3) & Clock (3) & Door (3) & Eyeglasses (3) & Faucet (3) & Knife (4) & Oven (3) & Refrigerator (3) & Storage (7) & \textbf{Average}\\
\hline
\multirow{2}{*}{Axis Ang}
 & ArtGS & 0.15 & 76.32 & 0.07 & 0.03 & \textbf{0.04} & 0.35 & 1.89 & \textbf{0.03} & \textbf{0.01} & 9.33 & 8.82\\
 & Ours & \textbf{0.02} & \textbf{0.00} & \textbf{0.05} & \textbf{0.01} & 0.05 & \textbf{0.32} & \textbf{0.30} & 0.09 & \textbf{0.01} & \textbf{0.03} & \textbf{0.09}\\
\cline{1-13}
\multirow{2}{*}{Axis Pos}
 & ArtGS & 0.56 & 3.32 & - & \textbf{0.00} & \textbf{0.00} & \textbf{0.00} & \textbf{0.37} & \textbf{0.01} & \textbf{0.00} & 1268.83 & 141.45\\
 & Ours & \textbf{0.01} & \textbf{0.00} & - & \textbf{0.00} & \textbf{0.00} & \textbf{0.00} & 0.80 & 0.03 & \textbf{0.00} & \textbf{0.00} & \textbf{0.09}\\
\cline{1-13}
\multirow{2}{*}{Geo Dist}
 & ArtGS & 12.07 & 82.85 & \textbf{0.00} & \textbf{0.04} & \textbf{0.03} & 0.35 & 27.05 & \textbf{0.04} & 0.04 & 5.85 & 12.83\\
 & Ours & \textbf{0.06} & \textbf{0.01} & \textbf{0.00} & \textbf{0.04} & 0.06 & \textbf{0.32} & \textbf{0.18} & 0.11 & \textbf{0.03} & \textbf{0.03} & \textbf{0.08}\\
\hline
\multirow{2}{*}{CD-s}
 & ArtGS & 0.96 & 0.45 & \textbf{2.98} & \textbf{0.23} & \textbf{0.07} & 0.51 & \textbf{0.84} & \textbf{0.85} & \textbf{0.90} & 1.27 & 0.90\\
 & Ours & \textbf{0.43} & \textbf{0.39} & 3.18 & 0.29 & \textbf{0.07} & \textbf{0.42} & 1.21 & 1.25 & 0.91 & \textbf{0.50} & \textbf{0.86}\\
\cline{1-13}
\multirow{2}{*}{CD-m}
 & ArtGS & 2.82 & 873.69 & \textbf{0.91} & 0.17 & 0.08 & 0.26 & 176.05 & 0.58 & \textbf{0.15} & 102.33 & 115.70\\
 & Ours & 0.15 & \textbf{0.07} & 5.83 & \textbf{0.10} & \textbf{0.03} & \textbf{0.04} & \textbf{0.10} & \textbf{0.51} & \textbf{0.15} & \textbf{0.10} & \textbf{0.71}\\
\cline{1-13}
\multirow{2}{*}{CD-w}
 & ArtGS & 1.70 & 1.35 & \textbf{2.09} & 0.34 & 0.10 & \textbf{0.37} & 1.45 & \textbf{1.03} & 1.05 & \textbf{1.74} & \textbf{1.12}\\
 & Ours & \textbf{0.76} & \textbf{0.45} & \textbf{2.09} & \textbf{0.28} & \textbf{0.09} & 0.38 & \textbf{1.23} & \textbf{1.03} & \textbf{1.04} & 3.92 & 1.13\\
\hline
\end{tabular}
\end{adjustbox}
\label{tab:gear}
\end{table*}

\begin{figure*}[h]
\centering
\includegraphics[width=0.9\linewidth]{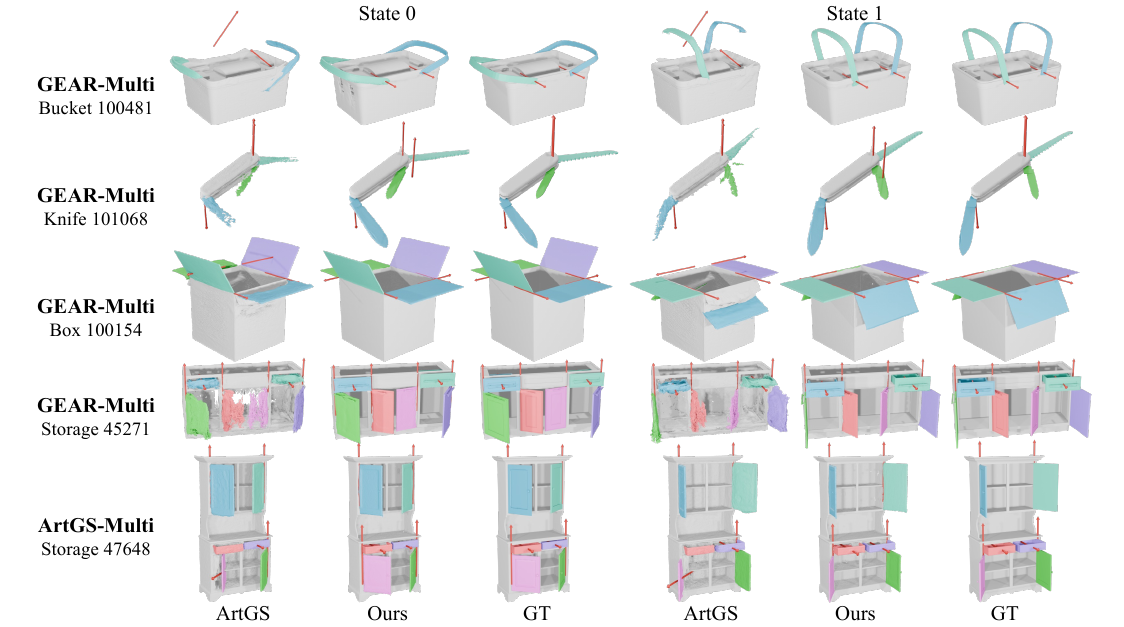}
\vspace{-0.5em}
\caption{Qualitative results on complex multi-joint articulated objects. Compared to ArtGS \cite{artgs}, Our method exhibits geometric reconstruction quality and more accurate motion parameter estimation.}
\label{fig:vis}
\vspace{-1.6em}
\end{figure*}

\cref{tab:paris} presents the results on the PARIS dataset. Overall, GEAR outperforms existing methods on most metrics, particularly achieving the lowest errors in motion parameter estimation in simulated objects. For geometric reconstruction, our method also attains high fidelity on both static and dynamic parts. While GEAR ranks second on the two real-world objects in PARIS, this marginal numerical gap primarily stems from ArtGS's dual-state canonical design aligning slightly better with these specific simple cases. Nevertheless, our visual fidelity remains strictly comparable, and as demonstrated across other datasets, ArtGS's design becomes notably less effective for highly articulated, multi-joint structures, where GEAR maintains robustness. Additional results on multi-joint real-world objects are detailed in the Supplementary Material.
\definecolor{rankone}{RGB}{220, 240, 220}
\definecolor{ranktwo}{RGB}{255, 249, 230}
\definecolor{rankthree}{RGB}{235, 245, 255}

\cref{tab:artgs} presents the evaluation results on the ArtGS-Multi dataset. From the table, we observe that GEAR achieves optimal average performance across all metrics, particularly in modeling the geometry of dynamic parts required for interactions. For certain objects, GEAR outperforms ArtGS, indicating its ability to handle multi-joint articulated object reconstruction.

\cref{tab:gear} presents the evaluation results on the GEAR-Multi dataset. Our method consistently outperforms existing baselines in both motion parameter estimation and geometric reconstruction. As object complexity increases---whether in the number of movable parts or the diversity of articulated structures---ArtGS begins to fail on certain categories (e.g., Bucket, Knife) and more intricate multi-joint objects (e.g., Storage). In contrast, GEAR demonstrates robustness across these challenging cases.
\cref{fig:vis} further provides qualitative examples from two datasets, showing that our method achieves part segmentations and motion parameters that more closely align with the ground-truth.

Moreover, \cref{fig:robot} showcases the use of GEAR for constructing digital-twin assets. We convert the reconstructed meshes and motion parameters into URDF files and deploy them in Omniverse Isaac Sim, enabling direct interaction with a robotic manipulator. These assets provide valuable articulated objects for the embodied AI community and help narrow the Sim2Real gap.
Additional visualizations in real-world and simulation, as well as failure cases, are provided in Supplementary  Sec. B and Sec. C, respectively.

\begin{table*}[t]
\centering
\caption{Ablation study on GEAR's components. We conduct experiments on two representative and highly complex 7-part objects (Storage\_47648 and Storage\_45271), lower ($\downarrow$) is better for all metrics.}
\vspace{-0.6em}

  \resizebox{\textwidth}{!}{
  \begin{tabular}{
    l c c c c c c
    c c c c c c
  }
    \toprule
    & \multicolumn{6}{c}{Storage\_47648} & \multicolumn{6}{c}{Storage\_45271} \\
    \cmidrule(lr){2-7} \cmidrule(lr){8-13}
    {Method} & {Axis Ang} & {Axis Pos} & {Geo Dist} & {CD-s} & {CD-m} & {CD-w} & {Axis Ang} & {Axis Pos} & {Geo Dist} & {CD-s} & {CD-m} & {CD-w} \\
    \midrule
    Full               & 0.07 & 0.01 & 0.10 & 0.61 & 0.32 & 0.60 & 0.03 & 0.00 & 0.03 & 0.50 & 0.10 & 3.92 \\
    2DGS$\to$3DGS      & 0.07 & 0.00 & 0.03 & 0.66 & 0.70 & 0.77 & 1.02 & 0.40 & 2.94 & 0.46 & 0.21 & 0.98 \\
    w/o self\_loss      & 22.67 & 0.94 & 30.38 & 1.48 & F & 2.42 & 35.83 & 475.96 & 24.91 & 11.78 & F & 6.16 \\
    w/o knn\_loss       & 0.05 & 0.00 & 0.09 & 0.52 & 20.93 & 0.59 & 0.02 & 0.00 & 0.02 & 0.49 & 32.38 & 0.90 \\
    staged optimization         & 15.28 & 0.66 & 13.36 & 2.07 & F & 1.43 & 29.61 & 1.29 & 7.04 & 11.72 & 24.34 & 1.05 \\
    joint optimization  & 27.79 & 0.17 & 16.22 & 0.81 & 35.11 & 0.59 & 21.71 & 1.96 & 13.76 & 6.18 & 260.04 & 0.91 \\
    \bottomrule
  \end{tabular}}
\label{tab:ablation}
\end{table*}

\subsection{Ablation Study}

We ablate key components of GEAR on two complex multi-joint objects (see \cref{tab:ablation}, where ``F'' denotes failure cases with vanishing parts). We evaluate: replacing 2DGS with 3DGS, removing the self-reconstruction loss (w/o self\_loss) or KNN clustering loss (w/o knn\_loss), and replacing our EM framework with staged or joint optimization.

\begin{figure*}[t]
\vspace{-0.6em}
  \centering
  \begin{minipage}{0.48\linewidth}
    \centering
    \includegraphics[width=\linewidth]{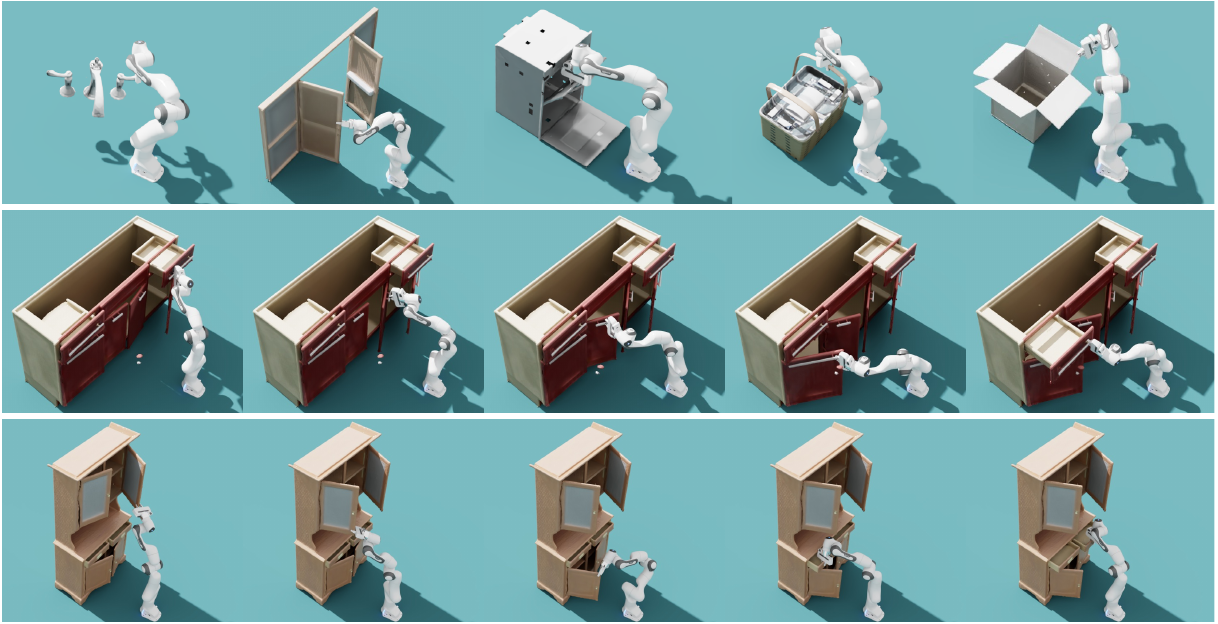}
    \caption{Interaction with assets generated by GEAR in Isaac Sim.}
    \label{fig:robot}
  \end{minipage}
  \hfill 
  \begin{minipage}{0.48\linewidth}
    \centering
    \includegraphics[width=\linewidth]{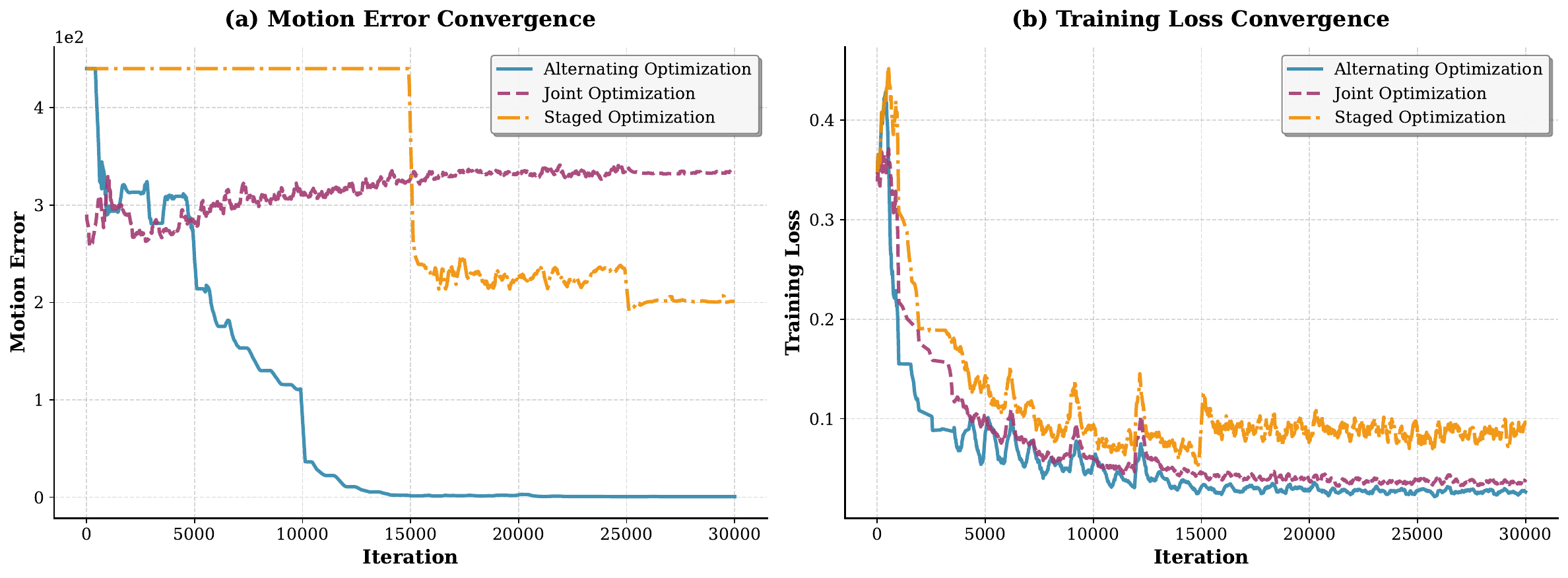}
    \caption{\textbf{Convergence Analysis on Storage\_45271.} (a) Sum of motion parameter errors (Axis Angle + Position + Geometry Distance) over iterations. (b) Total training loss over iterations. Our alternating optimization (blue) achieves the lowest error and loss, avoiding local minima traps and error accumulation.}
    \label{fig:convergence}
  \end{minipage}
  \vspace{-1.0em}
\end{figure*}

While 3DGS yields comparable overall accuracy, 2DGS provides better stability and lower reconstruction errors. For our single-state canonical field, the self-reconstruction loss is crucial to prevent overfitting to the target state. Additionally, the KNN clustering loss, though slightly restricting flexibility, effectively prevents mask drifting and significantly improves dynamic part segmentation.

To validate our EM formulation, we analyze the training dynamics on the challenging 7-part \textit{Storage\_45271} (\cref{fig:convergence}). Tracking the total motion error (\cref{fig:convergence}a), the joint optimization curve oscillates heavily without converging, empirically validating that error attribution ambiguity (\cref{subsec:em_optimization}) traps it in non-physical local minima. Meanwhile, staged optimization descends initially but stagnates, as its unidirectional pipeline allows early segmentation errors to irreversibly corrupt motion estimation.

In contrast, our alternating EM formulation effectively breaks this dependency loop. Its ``step-like'' error drops during E/M transitions prove that segmentation and motion estimation mutually guide each other towards stable convergence. Furthermore, our method achieves a significantly lower final training loss (\cref{fig:convergence}b). The minor periodic loss fluctuations are expected, resulting inherently from phase transitions and adaptive Gaussian densification.
\section{Conclusion}
\label{sec:conclusion}

In this paper, we presented GEAR, an EM-style alternating optimization framework for reconstructing articulated objects using Gaussian Splatting. By decomposing the tightly coupled geometry-motion optimization problem into an E-step and an M-step, and incorporating SAM-guided weak supervision, GEAR enables stable optimization without category-specific fine-tuning. Our framework achieves strong convergence on complex multi-joint objects under a coarse-to-fine initialization strategy. Extensive experiments demonstrate that GEAR delivers high-fidelity geometric reconstructions and highly accurate motion parameters, outperforming existing baselines. While our current method faces challenges with extreme articulations (e.g., 180-degree rotations) and transparent materials, future work will explore integrating motion priors and extending the representation for broader physical properties.

\noindent \textbf{Acknowledgement.} This work is partially supported by Beijing Municipal Natural Science Foundation Nos. L257009, L242025, and Natural Science Foundation of China under contracts Nos. 62495082, 62461160331, U21B2025.

{
    \small
    \bibliographystyle{ieeenat_fullname}
    \bibliography{main}
}
\appendix

\clearpage
\maketitlesupplementary

\setcounter{section}{0}
\renewcommand\thesection{\Alph{section}}
\renewcommand\thesubsection{\Alph{section}.\arabic{subsection}}
\renewcommand{\thefigure}{S\arabic{figure}}
\renewcommand{\thetable}{S\arabic{table}}
\setcounter{figure}{0}
\setcounter{table}{0}
\section*{Supplementary Material Overview}

The supplementary is organized as follows:~\cref{sec:impl_details} provides implementation specifics, algorithmic strategies, and hyperparameters.~\cref{sec:ext_exps} presents comprehensive quantitative and qualitative results, including runtime analysis, EM framework robustness, and detailed performances on benchmarks. Finally,~\cref{sec:limitations} discusses the current limitations of our method and potential future extensions.


\section{Implementation Details}
\label{sec:impl_details}

\begin{figure*}[b]
    \centering
    \includegraphics[width=\linewidth]{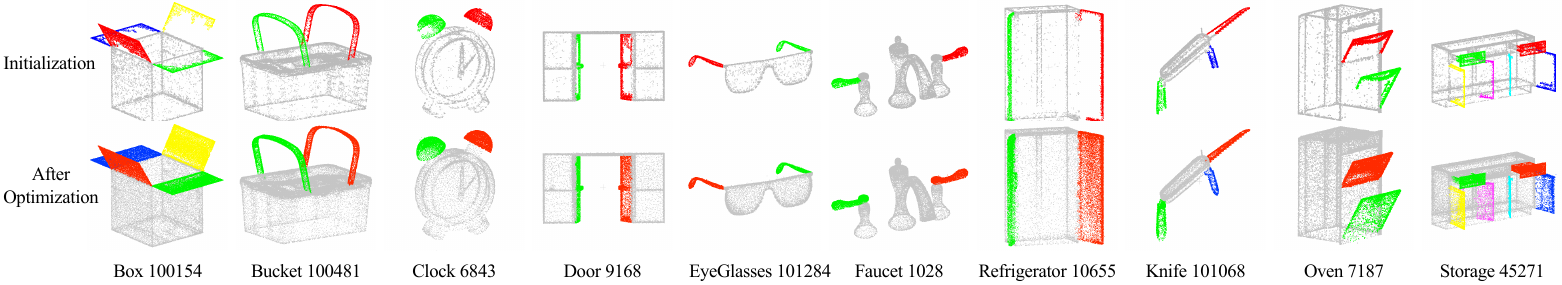} 
    \caption{\textbf{Visualization of Initialization and Convergence.} Top row: The coarse point clouds derived from the voxel-based initialization module. Colors indicate different initialized parts. Bottom row: The final high-fidelity  reconstruction after GEAR optimization. The comparison highlights that while the coarse module produces rough and discretized structures, it successfully captures the topological structure, guiding GEAR to converge to a photorealistic and geometrically accurate result.}
    \label{fig:coarse_vis}
\end{figure*}

In this section, we provide implementation specifics, algorithmic strategies, and detailed hyperparameters.
\subsection{Coarse Reconstruction Module}
To ensure consistent operations across different states, we establish a unified voxelization space. We adopt an adaptive voxel scale $v$ based on object complexity. Specifically, for simple single-joint objects, we use a coarser scale ($v=0.1$) to robustly cover and bound all potential dynamic candidates. For complex multi-joint objects, we employ a finer scale ($v=0.01$) combined with Top-K selection to isolate spatially adjacent small parts. Dynamic regions are extracted by subtracting the static intersection (processed with a morphological dilation) from the total voxel space.

Leveraging the prevalence of planar structures (e.g., doors, lids) in articulated objects, we apply a Structure-Aware Motion Initialization strategy. Extracted components with a high PCA-based aspect ratio (greater than 3.0) are classified as planes, allowing us to approximate the rotation axis via the intersection of fitted planes. For non-planar parts, we conservatively initialize with an identity matrix to avoid over-constraining the optimization. 

As demonstrated in Fig.~\ref{fig:coarse_vis}, the initialization prior acts as an anchor, guiding the subsequent GEAR optimization to refine geometry and motion, ultimately converging to a high-fidelity reconstruction.

\subsection{Part-Aware Optimization Refinements}
To handle the specific challenges posed by articulated objects with occlusions or thin structures, we introduce two targeted refinement strategies within the training loop:

\noindent\textbf{Class-Specific Voting for Prismatic Joints.} Prismatic joints (e.g., drawers, sliding doors) often exhibit limited visual changes compared to revolute joints and are frequently occluded by the static main body. During the SAM Mask Aggregation phase in the E-step, standard majority voting might allow the dominant static class to overwhelm these smaller dynamic regions. To mitigate this, we apply a voting boost factor $\gamma_{\text{prism}} = 4$ to the prismatic parts identified during optimization. Specifically, when aggregating SAM mask regions, votes cast for prismatic parts are weighted by $\gamma_{\text{prism}}$, ensuring that partially occluded drawers are correctly segmented and not absorbed by the static part.

\noindent\textbf{Adaptive Opacity Maintenance for Small Parts.} The standard 3D Gaussian Splatting pipeline periodically resets the opacity of all Gaussians to prevent local minima and encourage densification. However, for articulated parts with thin structures or small surface areas (e.g., handles, levers), the number of initialized Gaussians is often low. A global opacity reset can cause these valid but sparse Gaussians to be aggressively pruned. We implement a protective mechanism: before each opacity reset interval, we count the number of Gaussians assigned to each dynamic part. If a part contains fewer than $N_{\text{thresh}} = 2000$ points, we skip the opacity reset step for that specific part, ensuring its structural integrity is maintained throughout the optimization.

\subsection{Hyperparameters}
We list the key hyperparameters used in GEAR in \cref{tab:hyperparams}. The values kept consistent across all experiments unless otherwise stated.

\renewcommand{\thetable}{S\arabic{table}}
\setcounter{table}{0}

\begin{table}[h]
    \centering
    \small
    \caption{\textbf{Hyperparameters setting of GEAR.}}
    \resizebox{\linewidth}{!}{
    \begin{tabular}{lcc}
    \toprule
    \textbf{Category} & \textbf{Parameter} & \textbf{Value} \\
    \midrule
    \multirow{2}{*}{Initialization} 
        & Voxel Size $v$ & 0.1 / 0.01$^{*}$ \\
        & Morphological Dilation Radius & 1 \\
    \midrule
    \multirow{6}{*}{Training Loop} 
        & Total Iterations & 30,000 \\
        & E-step Interval & 500 \\
        & M-step Interval & 500 \\
        & KNN Gaussian Neighbors $\mathcal{N}$ & 3 \\
        & Joint Classification Iteration & 10,000 \\
        & Rotation Angle Threshold $\epsilon$ & $10^\circ$ \\
    \midrule
    \multirow{4}{*}{Loss Weights} 
        & D-SSIM Loss Weight $\lambda_\text{D-SSIM}$ & 0.2 \\
        & Depth Loss Weight $\lambda_\text{Depth}$ & 1.0 \\
        & Mask Cross-Entropy Weight $\lambda_\text{SAM}$ & 0.1 \\
        & KNN Consistency Weight $\lambda_\text{KNN}$ & 0.1 \\
    \bottomrule
    \end{tabular}
    }
    {\footnotesize $^{*}$ Adaptive: 0.1 for single-joint and 0.01 for multi-joint objects.}
    \label{tab:hyperparams}
\end{table}

\section{Extended Experiments and Analysis}
\label{sec:ext_exps}

In this section, we provide comprehensive quantitative and qualitative results to further validate the efficiency, robustness, and generalizability of GEAR.

\subsection{Computational Cost and Runtime Analysis}
\label{sec:runtime}

To evaluate the computational efficiency, we benchmark the runtime of GEAR against ArtGS~\cite{artgs}, on a single NVIDIA RTX 4090 GPU.

\noindent\textbf{Runtime.} 
Our full optimization schedule (30k iterations for initialization + 30k for alternating refinement) is designed to prioritize maximum stability for extremely complex multi-joint objects. As detailed in Tab.~\ref{tab:detailed_runtime}, the overall processing time scales with the complexity (i.e., the number of movable parts) of the object. While the complex \textit{Storage} object (7 parts) demands a longer convergence time (40 mins), simpler objects are typically reconstructed faster. Overall, the full pipeline averages approximately 34.0 minutes per object on the GEAR-Multi dataset.

\begin{table}[h]
\centering
\caption{\textbf{Runtime Breakdown on GEAR-Multi.} Time is measured in minutes. The total computational cost scales reasonably with the number of movable parts.}
\resizebox{\linewidth}{!}{
\begin{tabular}{l c c c c}
\toprule
\textbf{Object} & \textbf{Parts} & \textbf{Initialization} & \textbf{training} & \textbf{Total Time} \\
\midrule
Box & 5 & 13 & 35 & 48 \\
Bucket & 3 & 13 & 27 & 40 \\
Clock & 3 & 9 & 16 & 25 \\
Door & 3 & 12 & 17 & 29 \\
EyeGlasses & 3 & 10 & 24 & 34 \\
Faucet & 3 & 9 & 20 & 29 \\
Knife & 4 & 9 & 16 & 25 \\
Oven & 3 & 12 & 23 & 35 \\
Refrigerator & 3 & 13 & 22 & 35 \\
Storage & 7 & 14 & 26 & 40 \\
\midrule
\textbf{Average} & - & 11.4 & 22.6 & 34.0 \\
\bottomrule
\end{tabular}
}
\label{tab:detailed_runtime}
\end{table}

\noindent\textbf{Efficiency and Convergence.}
Importantly, our alternating EM framework intrinsically converges faster than joint optimization. To demonstrate its efficiency, we deploy a fast version, \textbf{Ours$^*$}, which halts optimization at 10k initialization iterations and 10k training iterations. As shown in Tab.~\ref{tab:fast_performance}, \textbf{Ours$^*$} requires only \textbf{11.4 minutes} on average---faster than ArtGS (14.0 minutes)---yet it still substantially outperforms ArtGS across most metrics. Furthermore, GEAR matches ArtGS in VRAM efficiency ($<$ 8GB), lowering the hardware requirements.

\begin{table}[h]
\centering
\caption{Performance of the Fast Version on GEAR-Multi (Average). Our fast version (\textbf{Ours$^*$}) outperforms ArtGS in runtime while delivering substantially higher accuracy.}
\resizebox{\columnwidth}{!}{
\begin{tabular}{l c | c c c | c c c}
\toprule
\textbf{Method} & \textbf{Time (min)}$\downarrow$ & \textbf{Axis Ang}$\downarrow$ & \textbf{Axis Pos}$\downarrow$ & \textbf{Geo Dist}$\downarrow$ & \textbf{CD-s}$\downarrow$ & \textbf{CD-m}$\downarrow$ & \textbf{CD-w}$\downarrow$ \\
\midrule
ArtGS~\cite{artgs} & 14.0 & 8.82 & 141.45 & 12.83 & 0.90 & 115.70 & 1.12 \\
\textbf{Ours$^*$ (Fast)} & \textbf{11.4} & 0.30 & \textbf{0.01} & 0.34 & 0.99 & 3.00 & \textbf{0.95} \\
Ours (Full) & 34.0 & \textbf{0.09} & 0.09 & \textbf{0.08} & \textbf{0.86} & \textbf{0.71} & 1.13 \\
\bottomrule
\end{tabular}
}
\label{tab:fast_performance}
\end{table}

\begin{table}[b]
\centering
\caption{\textbf{Alternating vs. Joint Optimization} on PARIS dataset. The results indicate that GEAR works reasonably well under joint optimization, while EM strategy behaves better.}
\resizebox{0.9\columnwidth}{!}{
\begin{tabular}{l c c c c c c}
\toprule
\textbf{Method} & \textbf{Axis Ang}$\downarrow$ & \textbf{Axis Pos}$\downarrow$ & \textbf{Geo Dist}$\downarrow$ & \textbf{CD-s}$\downarrow$ & \textbf{CD-m}$\downarrow$ & \textbf{CD-w}$\downarrow$ \\
\midrule
Ours (Joint) & 0.11 & \textbf{0.00} & 0.32 & 2.32 & 2.19 & 1.93 \\
Ours (EM) & \textbf{0.02} & \textbf{0.00} & \textbf{0.02} & \textbf{1.99} & \textbf{0.70} & \textbf{1.87} \\
\bottomrule
\end{tabular}
}
\label{tab:alt_vs_joint}
\end{table}

\begin{table}[b]
\centering
\caption{\textbf{Plugin Capability.} Applying EM framework (15k joint pre-training followed by 5k EM alternating refinement) to ArtGS~\cite{artgs} on the GEAR-Multi dataset reduces modeling errors.}
\resizebox{0.9\columnwidth}{!}{
\begin{tabular}{l c c c c c c}
\toprule
\textbf{Method} & \textbf{Axis Ang}$\downarrow$ & \textbf{Axis Pos}$\downarrow$ & \textbf{Geo Dist}$\downarrow$ & \textbf{CD-s}$\downarrow$ & \textbf{CD-m}$\downarrow$ & \textbf{CD-w}$\downarrow$ \\
\midrule
ArtGS & 8.82 & 141.45 & 12.83 & 0.90 & 115.70 & \textbf{1.12} \\
ArtGS + EM & \textbf{7.77} & \textbf{0.86} & \textbf{11.33} & \textbf{0.80} & \textbf{27.46} & 1.37 \\
\bottomrule
\end{tabular}
}
\label{tab:plugin_capability}
\end{table}

\subsection{Effectiveness of the EM Framework}
\label{sec:em_generality}

\begin{table*}[t]
\centering
\caption{\textbf{Robustness Analysis of the EM Alternating Interval.} Evaluated on two complex 7-part objects. GEAR maintains stable, high-quality convergence across a broad range of interval settings (50 to 2000). Degeneration only occurs when the interval is excessively large ($\ge 3000$), mimicking the failure mode of staged optimization.}
\resizebox{\textwidth}{!}{
\begin{tabular}{l c c c c c c c c c c c c}
\toprule
\multirow{2}{*}{\textbf{Method}} & \multicolumn{6}{c}{\textbf{Storage\_47648}} & \multicolumn{6}{c}{\textbf{Storage\_45271}} \\
\cmidrule(lr){2-7} \cmidrule(lr){8-13}
 & \textbf{Axis Ang}$\downarrow$ & \textbf{Axis Pos}$\downarrow$ & \textbf{Geo Dist}$\downarrow$ & \textbf{CD-s}$\downarrow$ & \textbf{CD-m}$\downarrow$ & \textbf{CD-w}$\downarrow$ & \textbf{Axis Ang}$\downarrow$ & \textbf{Axis Pos}$\downarrow$ & \textbf{Geo Dist}$\downarrow$ & \textbf{CD-s}$\downarrow$ & \textbf{CD-m}$\downarrow$ & \textbf{CD-w}$\downarrow$ \\
\midrule
ArtGS & 0.14 & 0.02 & 0.62 & 0.67 & 3.70 & 0.70 & 9.33 & 1268.83 & 5.85 & 1.27 & 102.33 & \textbf{1.74} \\
\midrule
Ours-50   & 0.06 & \textbf{0.01} & 0.13 & \textbf{0.53} & 0.23 & 0.60 & 0.03 & \textbf{0.00} & 0.07 & 0.49 & 0.17 & \textbf{0.89} \\
Ours-250  & 0.13 & \textbf{0.01} & 0.15 & 0.57 & \textbf{0.22} & \textbf{0.59} & \textbf{0.02} & \textbf{0.00} & 0.05 & 0.55 & 0.17 & 0.91 \\
Ours-500 (Default) & 0.07 & \textbf{0.01} & \textbf{0.10} & 0.61 & 0.32 & 0.60 & 0.03 & \textbf{0.00} & \textbf{0.03} & 0.50 & \textbf{0.10} & 3.92 \\
Ours-2000 & \textbf{0.05} & \textbf{0.01} & 0.12 & 0.54 & 0.26 & 0.60 & 0.03 & \textbf{0.00} & \textbf{0.03} & \textbf{0.43} & 0.16 & 0.91 \\
Ours-3000 & 0.19 & \textbf{0.01} & 0.24 & 0.55 & 0.24 & 0.60 & 13.17 & 3.02 & 9.54 & 1.00 & 73.78 & 0.94 \\
\bottomrule
\end{tabular}
}
\label{tab:em_interval_ablation}
\end{table*}

The contribution of GEAR lies in the EM-style alternating optimization framework. The ablation study in the main paper on complex multi-joint objects demonstrates that standard joint optimization struggles to converge. Here, we provide further insights into joint optimization on simpler tasks and the plugin generality of our EM framework.

\noindent\textbf{Joint Optimization Performance on Simple Objects.} 
To investigate whether the underlying GEAR representation (part-assigned Gaussians and dual-quaternion kinematics) inherently relies on the alternating strategy, we evaluate the joint optimization strategy on simpler, single-joint objects from the PARIS dataset. As shown in Tab.~\ref{tab:alt_vs_joint}, optimizing the GEAR representation jointly yields reasonable results on simple objects. 

This confirms that our fundamental geometric and motion representation is sound for basic tasks where error attribution ambiguity is relatively low. While joint optimization is applicable in these cases, the EM alternating strategy further reduces the errors (Angle error to $0.02$ and CD-m to 0.70). More importantly, as demonstrated in the main paper, when object complexity scales up to multiple joints, the EM framework becomes necessary to prevent the optimizer from getting trapped in local minima.

\noindent\textbf{Plugin Capability for Existing Methods.} 
To demonstrate that our EM-style formulation also serves as a generalizable optimization plugin, we integrate alternating strategy into ArtGS~\cite{artgs}. As shown in Tab.~\ref{tab:plugin_capability}, standard ArtGS struggles on GEAR-Multi dataset. To address this, we modify its training schedule: using its default joint optimization for the first 15k iterations, followed by our EM alternating refinement for the final 5k iterations (\textbf{ArtGS + EM}). 

This plug-and-play refinement substantially reduces errors: the Axis Pos error drops from 141.45 to 0.86, and the CD-m decreases from 115.70 to 27.46. This validates that our alternating paradigm is not only effective for training from scratch, but also a robust approach to recovering existing methods from local minima.

\subsection{Robustness of the EM Alternating Interval}
\label{sec:em_robustness}

To demonstrate that GEAR does not rely on exhaustive per-object hyperparameter tuning, we evaluate its sensitivity to the EM alternating interval (iterations per E/M-step) on two 7-part objects (\textit{Storage\_47648} and \textit{Storage\_45271}). 

As shown in Tab.~\ref{tab:em_interval_ablation}, GEAR stably maintains high-fidelity convergence across a wide range of intervals (50 to 2000 iterations), consistently outperforming the ArtGS baseline. Degeneration only occurs at excessively large intervals (e.g., 3000), where the process mimics a single-pass ``Staged Optimization," causing early segmentation errors to irrecoverably propagate. This confirms our framework's inherent stability for diverse articulated objects.

\subsection{Robustness to Imperfect 2D Mask}
\label{sec:sam_robustness}

\begin{figure}[h]
  \centering
  \includegraphics[width=0.9\linewidth]{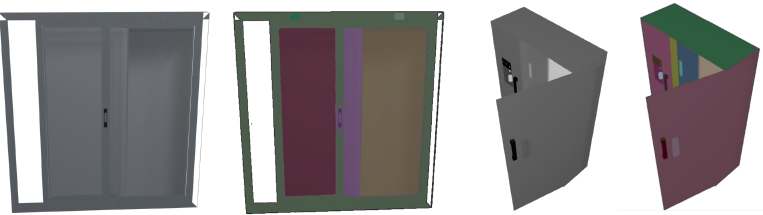}
  \caption{\textbf{Failure Cases of SAM.} Visualizations of flawed 2D masks, including over-segmentation (left) and under-segmentation (right), which serve as the challenging initial priors for our framework.}
  \label{fig:sam_robustness}
  \end{figure}

To evaluate the robustness of our framework against imperfect 2D priors, we analyze its performance under typical failure cases of SAM, such as over-segmentation (fragmenting parts) or under-segmentation (merging parts). As visualized in Fig.~\ref{fig:sam_robustness}, SAM produces flawed 2D priors. Nevertheless, as quantitatively validated in Tab.~\ref{tab:sam_robustness}, GEAR effectively filters out this segmentation noise, yielding highly accurate geometric and motion estimates on these failure-prone objects.

\begin{table}[h]
  \centering
  \caption{\textbf{Performance under Flawed 2D Priors.} GEAR maintains high accuracy despite being initialized with the erroneous masks shown in~\cref{fig:sam_robustness}.}  
  \resizebox{\columnwidth}{!}{
    \begin{tabular}{l | c c c c c c}
      \toprule
      \textbf{Object} & \textbf{Axis Ang}$\downarrow$ & \textbf{Axis Pos}$\downarrow$ & \textbf{Geo Dist}$\downarrow$ & \textbf{CD-s}$\downarrow$ & \textbf{CD-m}$\downarrow$ & \textbf{CD-w}$\downarrow$ \\
      \midrule
      Window\_102985 & 0.06 & - & 0.00 & 0.77 & 2.10 & 0.68 \\
      Refrigerator\_10685 & 0.00 & 0.00 & 0.02 & 0.52 & 0.10 & 0.53 \\
      \bottomrule
    \end{tabular}
  }
  \label{tab:sam_robustness}
\end{table}

\begin{figure*}[b]
    \centering
    \vspace{-2em}
    \includegraphics[width=0.75\linewidth]{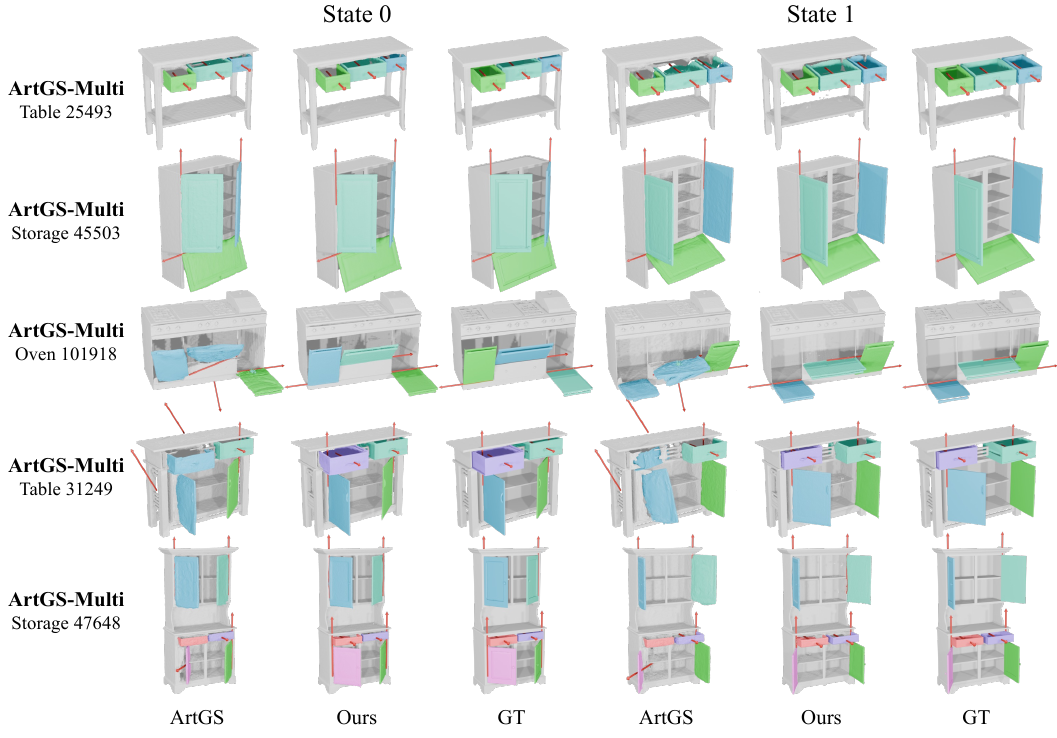} 
    \caption{\textbf{Qualitative comparison on the ArtGS-Multi dataset.} We visualize all reconstruction results across two states. The columns show the results from the ArtGS baseline, Ours, and the Ground Truth (GT). As highlighted in the visualizations (e.g., \textit{Oven 101918} and \textit{Table 31249}), ArtGS struggles to distinguish spatially adjacent parts, leading to fused segmentation. In contrast, GEAR accurately disentangles these components and estimates precise motion axes.}
    \label{fig:sup-artgs}
    \vspace{-0.6em}
\end{figure*}

\begin{table*}[t]
    \centering
    \caption{ Detailed quantitative comparison on the PARIS dataset, averaged over 10 trials for both simulated and real objects with higher visibility. 
  Lower ($\downarrow$) is better for all metrics. 
  The best (\textbf{Bold} with \colorbox{rankone!90}{Top-1}) and second-best (\colorbox{ranktwo!90}{Top-2}) results are highlighted. 
  Baseline results for Ditto, PARIS, DTA, and ArtGS are sourced from ArtGS, objects with * are seen categories trained in Ditto.}
    \label{tab:paris_full}
    \renewcommand{\arraystretch}{1.1}
    \setlength{\tabcolsep}{2.5pt}
    \begin{adjustbox}{width=0.75\textwidth}
    \begin{tabular}{c l | c c c c c c c c c c c | c c c}
    \hline
     & & \multicolumn{11}{c|}{Simulation} & \multicolumn{3}{c}{Real} \\
    \cline{3-13} \cline{14-16}
     & & Foldchair & Fridge & Laptop* & Oven* & Scissor & Stapler & USB & Washer & Blade & Storage* & Average & Fridge & Storage* & Average \\
    \hline
    \multirow{6}{*}{Axis Ang}
     & Ditto & 89.35 & 89.30 & 3.12 & 0.96 & 4.50 & 89.86 & 89.77 & 89.51 & 79.54 & 6.32 & 46.22 & \cellcolor{ranktwo!90}1.71 & 5.88 & 3.80 \\
     & PARIS & 7.90 & 9.19 & \cellcolor{ranktwo!90}0.02 & \cellcolor{ranktwo!90}0.04 & 3.92 & 0.73 & 0.13 & 25.18 & 15.18 & 0.03 & 6.23 & \cellcolor{rankone!90}\textbf{1.64} & 43.13 & 22.39 \\
     & DTA & 0.03 & 0.09 & 0.07 & 0.22 & 0.10 & \cellcolor{ranktwo!90}0.07 & 0.11 & 0.36 & 0.20 & 0.09 & \cellcolor{ranktwo!90}0.13 & 2.08 & 13.64 & 7.86 \\
     & ArtGS & \cellcolor{ranktwo!90}0.01 & \cellcolor{ranktwo!90}0.03 & \cellcolor{rankone!90}\textbf{0.01} & \cellcolor{rankone!90}\textbf{0.01} & \cellcolor{ranktwo!90}0.05 & \cellcolor{rankone!90}\textbf{0.01} & \cellcolor{ranktwo!90}0.04 & \cellcolor{rankone!90}\textbf{0.02} & \cellcolor{rankone!90}\textbf{0.03} &\cellcolor{ranktwo!90} 0.02 & \cellcolor{rankone!90}\textbf{0.02} & 2.09 & \cellcolor{rankone!90}\textbf{3.47} & \cellcolor{rankone!90}\textbf{2.78} \\
     & Ours & \cellcolor{rankone!90}\textbf{0.00} & \cellcolor{rankone!90}\textbf{0.01} & \cellcolor{ranktwo!90}0.02 & \cellcolor{rankone!90}\textbf{0.01} & \cellcolor{rankone!90}\textbf{0.00} & \cellcolor{rankone!90}\textbf{0.01} & \cellcolor{rankone!90}\textbf{0.01} & \cellcolor{ranktwo!90}0.05 & \cellcolor{ranktwo!90}0.09 & \cellcolor{rankone!90}\textbf{0.00} & \cellcolor{rankone!90}\textbf{0.02} & 2.16 & \cellcolor{ranktwo!90}4.85 & \cellcolor{ranktwo!90}3.50 \\
    \cline{1-16}
    \multirow{6}{*}{Axis Pos}
     & Ditto & 3.77 & 1.02 & \cellcolor{ranktwo!90}0.01 & 0.13 & 5.70 & 0.20 & 5.41 & 0.66 & - & - & 2.11 & 1.84 & - & 1.84 \\
     & PARIS & 0.37 & 0.30 & 0.02 & \cellcolor{rankone!90}\textbf{0.00} & 1.52 & 2.26 & \cellcolor{ranktwo!90}2.37 & 1.50 & - & - & 1.04 & \cellcolor{rankone!90}\textbf{0.34} & - & \cellcolor{rankone!90}\textbf{0.34} \\
     & DTA & \cellcolor{ranktwo!90}0.01 & \cellcolor{ranktwo!90}0.01 & \cellcolor{ranktwo!90}0.01 & \cellcolor{ranktwo!90}0.01 & \cellcolor{ranktwo!90}0.02 & 0.02 & \cellcolor{rankone!90}\textbf{0.00} & \cellcolor{ranktwo!90}0.05 & - & - & \cellcolor{ranktwo!90}0.02 & 0.59 & - & 0.59 \\
     & ArtGS & \cellcolor{rankone!90}\textbf{0.00} & \cellcolor{rankone!90}\textbf{0.00} & \cellcolor{ranktwo!90}0.01 & \cellcolor{rankone!90}\textbf{0.00} & \cellcolor{rankone!90}\textbf{0.00} & \cellcolor{ranktwo!90}0.01 & \cellcolor{rankone!90}\textbf{0.00} & \cellcolor{rankone!90}\textbf{0.00} & - & - & \cellcolor{rankone!90}\textbf{0.00} & 0.47 & - & 0.47 \\
     & Ours & \cellcolor{rankone!90}\textbf{0.00} & \cellcolor{rankone!90}\textbf{0.00} & \cellcolor{rankone!90}\textbf{0.00} & \cellcolor{rankone!90}\textbf{0.00} & \cellcolor{rankone!90}\textbf{0.00} & \cellcolor{rankone!90}\textbf{0.00} & \cellcolor{rankone!90}\textbf{0.00} & \cellcolor{rankone!90}\textbf{0.00} & - & - & \cellcolor{rankone!90}\textbf{0.00} & \cellcolor{ranktwo!90}0.38 & - & \cellcolor{ranktwo!90}0.38 \\
    \cline{1-16}
    \multirow{6}{*}{Geo Dist}
     & Ditto & 99.36  & F     & 5.18 & 2.09 & 19.28  & 56.61 & 80.60 & 55.72 & F    & 0.09 & 39.87 & 8.43 & 0.38 & 4.41 \\
     & PARIS & 131.82 & 24.64 & 3.03 & \cellcolor{ranktwo!90}0.04 & 120.61 & 10.71 & 64.91 & 60.62 & \cellcolor{ranktwo!90}0.54 & \cellcolor{ranktwo!90}0.14 & 41.71 & 2.16 & 0.56 & 1.36 \\
     & DTA   & 0.10   & 0.12  & 0.11 & 0.12 & 0.37   & 0.08  & 0.15  & \cellcolor{ranktwo!90}0.28  & \cellcolor{rankone!90}\textbf{0.00} & \cellcolor{rankone!90}\textbf{0.00} & \cellcolor{ranktwo!90}0.13  & \cellcolor{rankone!90}\textbf{1.85} & 0.14 & \cellcolor{ranktwo!90}1.00 \\
     & ArtGS & \cellcolor{rankone!90}\textbf{0.03}   & \cellcolor{ranktwo!90}0.04  & \cellcolor{ranktwo!90}0.02 & \cellcolor{rankone!90}\textbf{0.02} & \cellcolor{ranktwo!90}0.04   & \cellcolor{rankone!90}\textbf{0.01}  & \cellcolor{ranktwo!90}0.03  & \cellcolor{rankone!90}\textbf{0.03}  & \cellcolor{rankone!90}\textbf{0.00} & \cellcolor{rankone!90}\textbf{0.00} & \cellcolor{rankone!90}\textbf{0.02}  & \cellcolor{ranktwo!90}1.94 & \cellcolor{rankone!90}\textbf{0.04} & \cellcolor{rankone!90}\textbf{0.99} \\
     & Ours  & \cellcolor{ranktwo!90}0.04   & \cellcolor{rankone!90}\textbf{0.01}  & \cellcolor{rankone!90}\textbf{0.01} & \cellcolor{rankone!90}\textbf{0.02} & \cellcolor{rankone!90}\textbf{0.01}   & \cellcolor{ranktwo!90}0.02  & \cellcolor{rankone!90}\textbf{0.01}  & \cellcolor{rankone!90}\textbf{0.03}  & \cellcolor{rankone!90}\textbf{0.00} & \cellcolor{rankone!90}\textbf{0.00} & \cellcolor{rankone!90}\textbf{0.02}  & 2.17 & \cellcolor{ranktwo!90}0.06 & 1.18 \\
    \hline
    \multirow{6}{*}{CD-s}
     & Ditto & 33.79 & 3.05 & \cellcolor{rankone!90}\textbf{0.25} & \cellcolor{rankone!90}\textbf{2.52}  & 39.07 & 41.64 & 2.64 & 10.32 & 46.90 & 9.18  & 18.94 & 47.01 & 16.09 & 31.55 \\
     & PARIS & 9.12  & 3.73 & 0.45 & 12.85 & 1.83  & \cellcolor{rankone!90}\textbf{1.96}  & 2.58 & 25.19 & 1.33  & 12.80 & 7.18  & 42.57 & 54.54 & 48.56 \\
     & DTA   & \cellcolor{rankone!90}\textbf{0.18}  & 0.62 & \cellcolor{ranktwo!90}0.30 & 4.60  & 3.55  & \cellcolor{ranktwo!90}2.91  & \cellcolor{ranktwo!90}2.32 & \cellcolor{rankone!90}\textbf{4.56}  & \cellcolor{ranktwo!90}0.55  & \cellcolor{ranktwo!90}4.90  & \cellcolor{ranktwo!90}2.45  & 2.36  & 10.98 & 6.67 \\
     & ArtGS & 0.26  & \cellcolor{ranktwo!90}0.52 & 0.63 & 3.88  & \cellcolor{ranktwo!90}0.61  & 3.83  & \cellcolor{rankone!90}\textbf{2.25} & 6.43  & \cellcolor{rankone!90}\textbf{0.54}  & 7.31  & 2.63  & \cellcolor{ranktwo!90}1.64  & \cellcolor{rankone!90}\textbf{2.93}  & \cellcolor{rankone!90}\textbf{2.29} \\
     & Ours  & \cellcolor{ranktwo!90}0.20  & \cellcolor{rankone!90}\textbf{0.44} & 0.53 & \cellcolor{ranktwo!90}2.73  & \cellcolor{rankone!90}\textbf{0.44}  & 3.89  & 2.72 & \cellcolor{ranktwo!90}5.28  & 0.67  & \cellcolor{rankone!90}\textbf{2.98}  & \cellcolor{rankone!90}\textbf{1.99}  & \cellcolor{rankone!90}\textbf{1.50}  & \cellcolor{ranktwo!90}3.70  & \cellcolor{ranktwo!90}2.60 \\
    \cline{1-16}
    \multirow{6}{*}{CD-m}
     & Ditto & 141.11 & 0.99 & 0.19 & 0.94  & 20.68 & 31.21 & 15.88 & 12.89  & 195.93 & 2.20  & 42.20 & 50.60 & 20.35  & 35.48 \\
     & PARIS & 8.79  & 7.76  & 0.49 & 28.51 & 46.69 & 19.36 & 5.53  & 178.39 & 25.29  & 76.75 & 39.76 & 45.66 & 864.82 & 455.24 \\
     & DTA   & \cellcolor{ranktwo!90}0.15  & 0.27  & \cellcolor{ranktwo!90}0.13 & \cellcolor{ranktwo!90}0.44  & 10.11 & 1.13  & 1.47  & \cellcolor{ranktwo!90}0.45   & 2.05   & \cellcolor{rankone!90}\textbf{0.36}  & 1.66  & 1.12  & 30.78  & 15.95 \\
     & ArtGS & 0.54  & \cellcolor{ranktwo!90}0.21  & \cellcolor{ranktwo!90}0.13 & 0.89  & \cellcolor{ranktwo!90}0.64  & \cellcolor{rankone!90}\textbf{0.52}  & \cellcolor{rankone!90}\textbf{1.22}  & \cellcolor{ranktwo!90}0.45   & \cellcolor{rankone!90}\textbf{1.12}   & \cellcolor{ranktwo!90}1.02  & \cellcolor{rankone!90}\textbf{0.67}  & \cellcolor{rankone!90}\textbf{0.66}  & \cellcolor{rankone!90}\textbf{6.28}   & \cellcolor{rankone!90}\textbf{3.47} \\
     & Ours  & \cellcolor{rankone!90}\textbf{0.12}  & \cellcolor{rankone!90}\textbf{0.19}  & \cellcolor{rankone!90}\textbf{0.11} & \cellcolor{rankone!90}\textbf{0.26}  & \cellcolor{rankone!90}\textbf{0.39}  & \cellcolor{ranktwo!90}0.56  & \cellcolor{ranktwo!90}1.39  & \cellcolor{rankone!90}\textbf{0.05}   & \cellcolor{ranktwo!90}1.89   & 2.02  & \cellcolor{ranktwo!90}0.70  & \cellcolor{ranktwo!90}0.69  & \cellcolor{ranktwo!90}16.01  & \cellcolor{ranktwo!90}8.35 \\
    \cline{1-16}
    \multirow{6}{*}{CD-w}
     & Ditto & 6.80 & 2.16 & \cellcolor{rankone!90}\textbf{0.31} & \cellcolor{ranktwo!90}2.51 & 1.70  & \cellcolor{ranktwo!90}2.38 & 2.09 & 7.29  & 42.04 & \cellcolor{ranktwo!90}3.91 & 7.12 & 6.50  & 14.08 & 10.29 \\
     & PARIS & 1.90 & 2.53 & 0.50 & \cellcolor{rankone!90}\textbf{1.94} & 10.20 & 6.30 & 2.31 & 24.71 & \cellcolor{ranktwo!90}0.44  & \cellcolor{rankone!90}\textbf{3.84} & 5.47 & 22.98 & 63.35 & 43.17 \\
     & DTA   & \cellcolor{ranktwo!90}0.27 & 0.70 & \cellcolor{ranktwo!90}0.32 & 4.24 & \cellcolor{rankone!90}\textbf{0.41}  & \cellcolor{rankone!90}\textbf{1.92} & \cellcolor{rankone!90}\textbf{1.17} & \cellcolor{rankone!90}\textbf{4.48}  & \cellcolor{rankone!90}\textbf{0.36}  & 3.99 & \cellcolor{rankone!90}\textbf{1.79} & 2.08  & 8.98  & 5.53 \\
     & ArtGS & 0.43 & \cellcolor{ranktwo!90}0.58 & 0.50 & 3.58 & 0.67  & 2.63 & \cellcolor{ranktwo!90}1.28 & 5.99  & 0.61  & 5.21 & 2.15 & \cellcolor{ranktwo!90}1.29  & \cellcolor{rankone!90}\textbf{3.23}  & \cellcolor{rankone!90}\textbf{2.26} \\
     & Ours  & \cellcolor{rankone!90}\textbf{0.23} & \cellcolor{rankone!90}\textbf{0.56} & 0.47 & 2.85 & \cellcolor{ranktwo!90}0.47  & 2.59 & 1.61 & \cellcolor{ranktwo!90}5.32  & 0.61  & 4.01 & \cellcolor{ranktwo!90}1.87 & \cellcolor{rankone!90}\textbf{1.00}  & \cellcolor{ranktwo!90}4.99  & \cellcolor{ranktwo!90}2.99 \\
    \hline
    \end{tabular}
    \end{adjustbox}
\end{table*}

\subsection{Detailed Results on PARIS Dataset}
Here, we present the complete per-object breakdown in Tab.~\ref{tab:paris_full}, which covers 10 synthetic objects and 2 real-world objects. As shown in the table, GEAR achieves the highest accuracy in motion parameter estimation across the majority of simulated objects.

\begin{figure*}[t]
    \centering
    \includegraphics[width=1\linewidth]{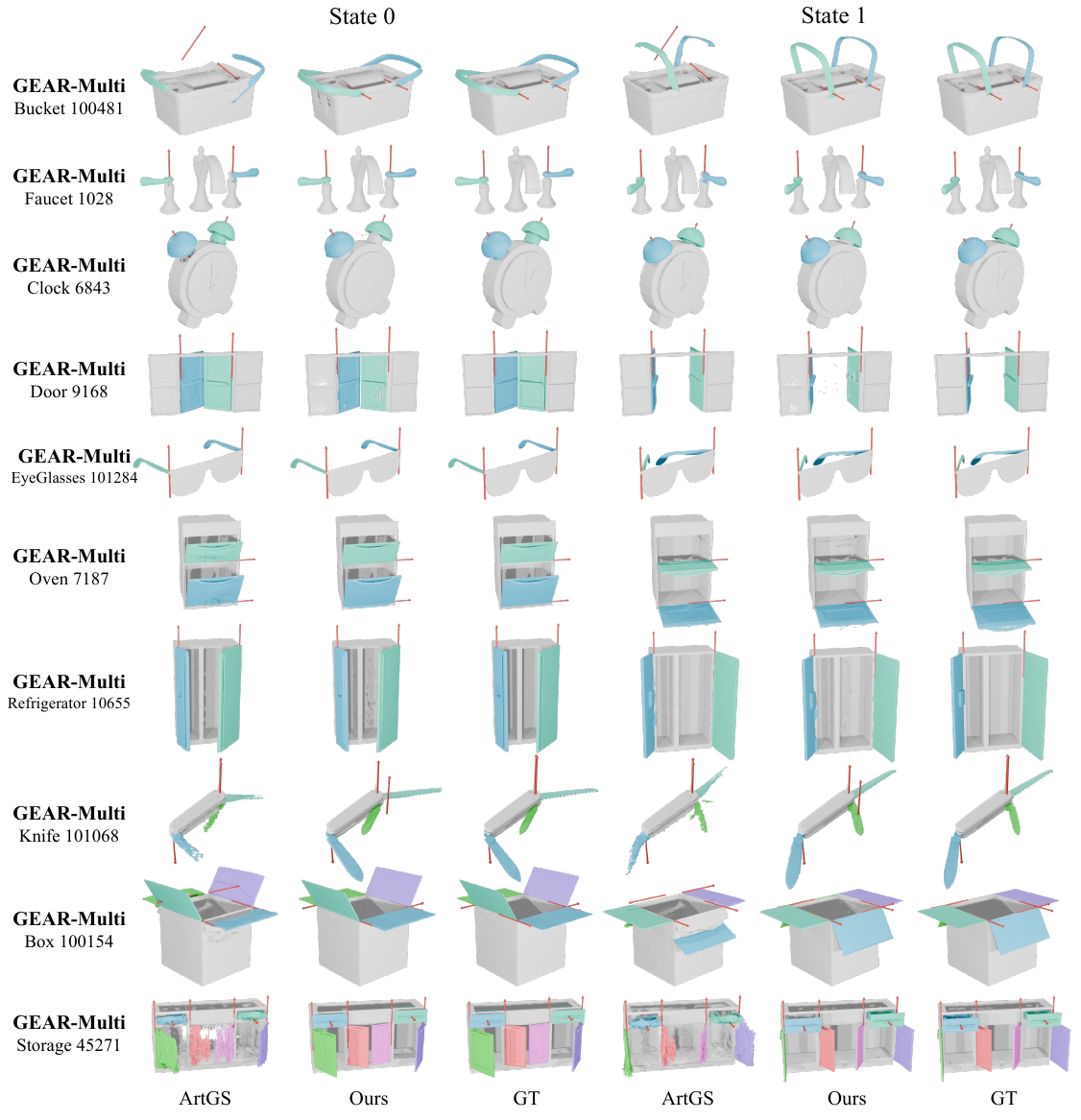} 
    \vspace{-0.0em}
    \caption{\textbf{Qualitative comparison on the GEAR-Multi dataset.} We present results on all 10 objects with diverse categories and topological structures. GEAR demonstrates robustness in handling objects with thin structures and small movable parts (e.g.,  \textit{Bucket 100481}, \textit{Knife 101068}), whereas ArtGS often fails to capture these fine-grained geometries or generates noisy artifacts.}
    \label{fig:sup-gear}
\end{figure*}

\subsection{Reconstruction Results} 

\begin{figure*}[t]
    \centering
    \includegraphics[width=0.9\linewidth]{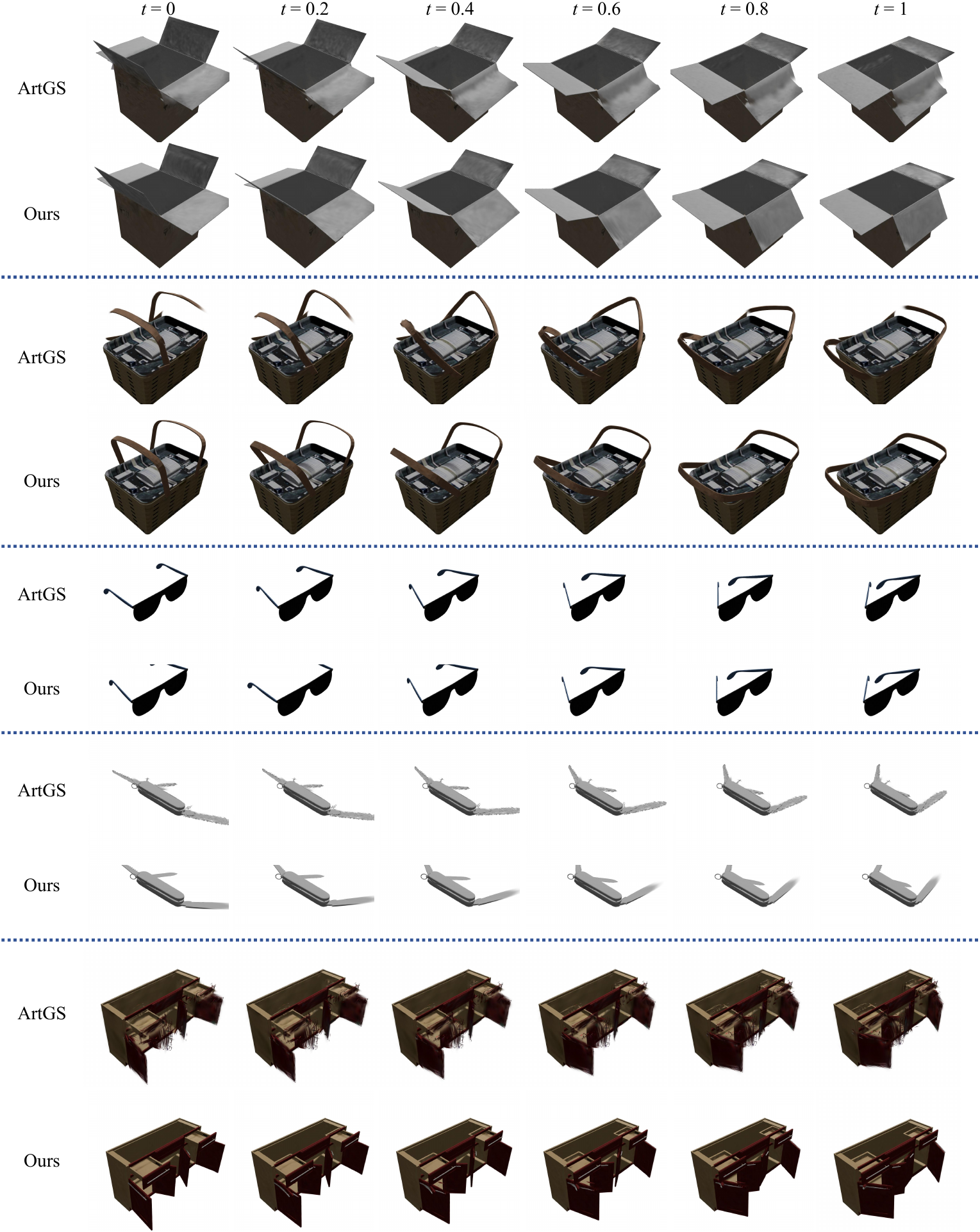} 
    \caption{\textbf{Novel Joint Motion Synthesis.} We render the articulated object at intermediate states $t=\{0, 0.2, 0.4, 0.6, 0.8, 1\}$. GEAR (Bottom) preserves the geometry of the door and the contents inside the fridge consistently throughout the motion trajectory, whereas the baseline (Top) suffers from geometric distortion and blurring.}
    \label{fig:interpolation}
\end{figure*}

\begin{figure*}[t]
    \centering
    \includegraphics[width=0.9\linewidth]{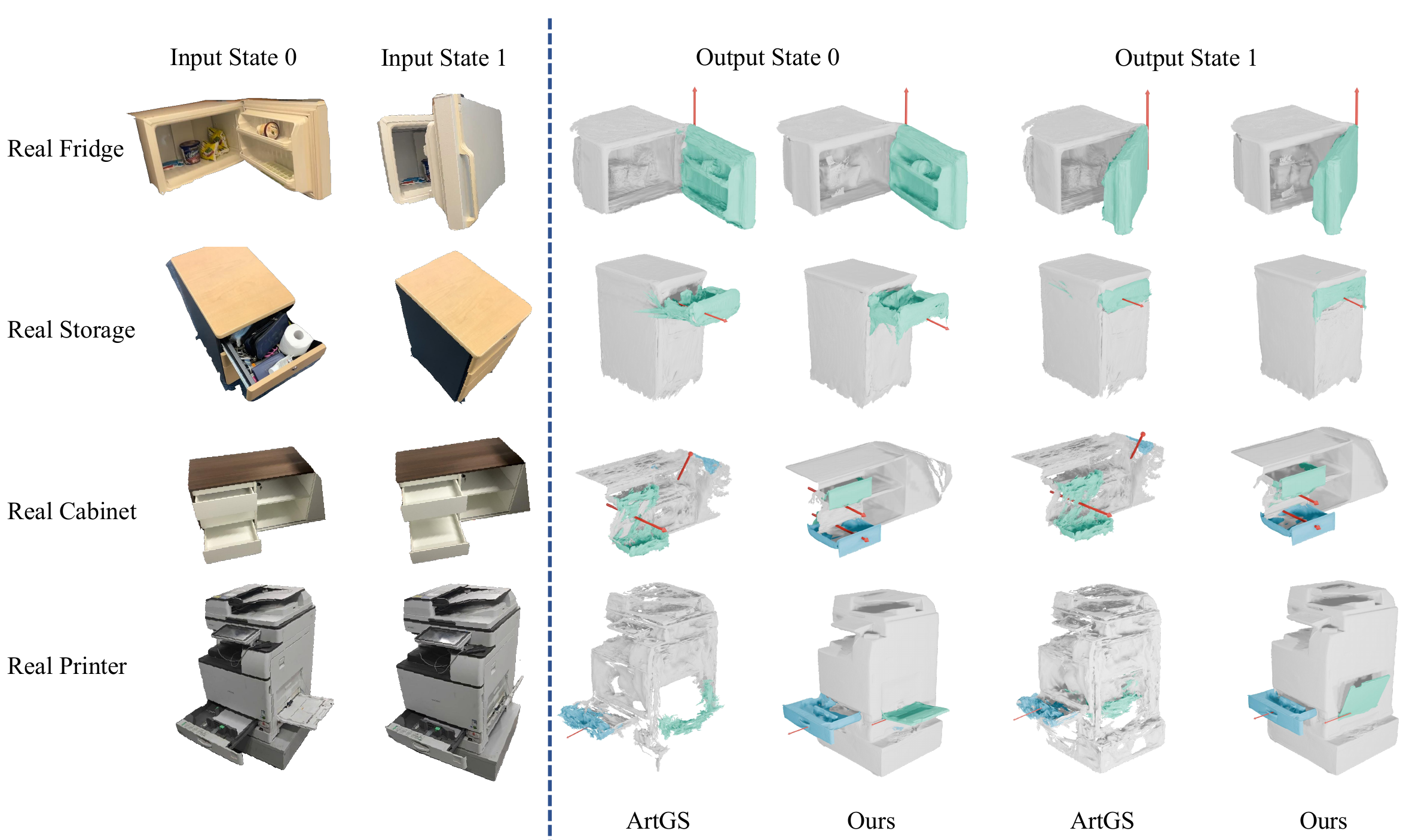}
    \caption{\textbf{Qualitative results on real-world objects.} The top two rows display single-joint objects from the PARIS dataset. The third row shows a multi-joint cabinet from the ArtGS dataset, and the bottom row depicts a printer captured in our lab. While ArtGS achieves comparable results on simple single-joint objects, it fails to reconstruct the structural details of complex multi-joint objects, leading to noisy artifacts. GEAR consistently delivers high-fidelity reconstruction and accurate part segmentation across all sources.}
    \label{fig:real_objects}
\end{figure*}

\begin{figure}[b]
    \centering
    \includegraphics[width=0.65\linewidth]{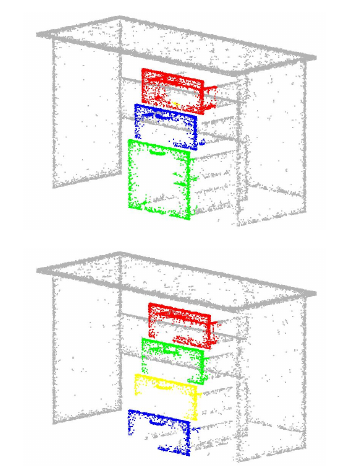} 
    \caption{\textbf{Spatial Ambiguity in Initialization.} Top: When two movable parts are spatially adjacent with no gap (parallel drawers), our coarse module may fail to separate them, initializing them as a single part. Bottom: When parts are spatially staggered (offset drawers), the module successfully identifies distinct components.}
    \label{fig:fail_spatial}
\end{figure}

We provide a comprehensive qualitative comparison between GEAR and the baseline ArtGS~\cite{artgs} across diverse objects from both ArtGS-Multi and GEAR-Multi datasets. As visualized in Fig.~\ref{fig:sup-artgs} and Fig.~\ref{fig:sup-gear}, GEAR successfully disentangles tightly packed adjacent parts and preserves high-frequency geometric details for thin structures (e.g., handles, knife blades), whereas the baseline often generates noisy artifacts or fused geometries. 

To validate the physical correctness of our estimated motion parameters, we render the articulated objects at intermediate continuous states. As shown in Fig.~\ref{fig:interpolation},  GEAR preserves the geometry of moving parts seamlessly throughout the entire motion trajectory without distortion.

\subsection{Performance on Real-World Objects}

To assess robustness against sensor noise and uncontrolled lighting, we evaluate GEAR on real-world objects. As visualized in Fig.~\ref{fig:real_objects}, while the baseline produces reasonable reconstructions on simple single-joint objects, it exhibits significant performance degradation on complex multi-joint inputs (e.g., Real Cabinet and Real Printer), resulting in severe floating artifacts and incorrect part assignments. In contrast, GEAR preserves the integrity of planar surfaces and accurately disentangles coupled motion parts under real-world conditions.

\section{Limitations}
\label{sec:limitations}

While GEAR demonstrates robust performance on a wide range of complex articulated objects, it is subject to certain limitations that present exciting avenues for future research.

\subsection{Current Limitations}

\noindent\textbf{Spatial Ambiguity in Coarse Initialization.}
Our coarse reconstruction module relies on Connected Component Analysis (CCA) of dynamic voxels to separate movable parts. This approach assumes that different movable parts are spatially disjoint in 3D space. However, when multiple movable parts are tightly packed or perfectly adjacent (e.g., side-by-side cabinet doors with negligible gaps), the dilated voxel grids may merge into a single connected component. 

As illustrated in Fig.~\ref{fig:fail_spatial}, we present a failure case with a 4-door cabinet where the two bottom doors are flush against each other. The initialization module incorrectly merges them into a single dynamic group. In contrast, we show a success case with the same object where the drawers are open in a \textit{staggered} configuration. The spatial offset allows the CCA to correctly identify them as distinct parts. This indicates that our current initialization requires a minimal spatial separation between moving parts.

\noindent\textbf{Misclassification in Extreme Articulation.}
GEAR relies on visual correspondence to estimate motion parameters. For extreme articulations, particularly 180-degree rotations, the visual overlap between the canonical state ($180^\circ$) and the target state ($0^\circ$) is minimal, and the geometric displacement is extremely large. In such scenarios, the optimization landscape is highly non-convex. We observe that the model may fall into a local minimum where it interprets the large displacement as a linear translation rather than a rotation.

\begin{figure}[h]
    \centering
    \includegraphics[width=0.9\linewidth]{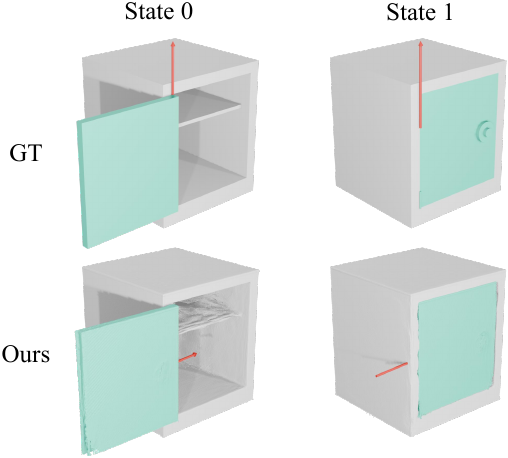}
    \caption{\textbf{Misclassification in Extreme Motion.} For a safe box door rotating 180 degrees, the large displacement and lack of intermediate visual cues lead the model to misclassify the revolute joint as a prismatic joint (visualized by the incorrect linear motion arrow).}
    \label{fig:fail_motion}
\end{figure}

Fig.~\ref{fig:fail_motion} demonstrates this on a \textit{Safe Box} object with a single revolute door opening 180 degrees. The model incorrectly identifies the joint as a prismatic joint, predicting a large translation vector instead of the correct rotation axis and angle. Integrating kinematic priors or trajectory constraints could be a potential solution.

\noindent\textbf{Material Constraints.}
Finally, our method inherits the inherent limitations of the Gaussian Splatting representation. GEAR struggles to reconstruct articulated objects made of transparent (e.g., glass cabinets) or highly reflective materials. The view-dependent effects of such materials are difficult to model with standard spherical harmonics, leading to noisy geometry or ``holes" in the reconstruction. Since GEAR uses geometric cues (depth and occupancy) for part segmentation, these rendering artifacts can propagate errors into the segmentation and motion estimation pipeline.

\subsection{Future Extensions}

To address these limitations and broaden GEAR's applicability, future work can explore several promising directions. First, integrating generative priors (e.g., diffusion models) could facilitate in-the-wild reconstruction from sparse observations by hallucinating unobserved geometry. Second, to resolve misclassifications in extreme motions (as shown in Fig.~\ref{fig:fail_motion}), incorporating 4DGS temporal modeling from continuous video sequences would allow the framework to explicitly track non-linear trajectories and avoid local minima. Finally, evolving our formulation into neural deformation fields would enable the simultaneous modeling of rigid part articulation and localized non-rigid dynamics for flexible objects.
\end{document}